\pdfoutput=1
\documentclass[11pt]{article}

\usepackage[final]{acl}
\usepackage{float}
\usepackage{times}
\usepackage{latexsym}
\usepackage{amsmath}
\usepackage{booktabs} 
\usepackage{tcolorbox}
\usepackage{multicol}
\usepackage{fancyvrb}
\usepackage{amsfonts}
\usepackage{enumitem}
\usepackage{afterpage}
\usepackage[T1]{fontenc}
\usepackage[utf8]{inputenc}

\usepackage{microtype}
\usepackage{multirow}
\usepackage{booktabs}
\usepackage{tabularx}
\usepackage{makecell}

\usepackage{graphicx}

\title{How Does Controllability Emerge In Language Models During Pretraining?}

\author{
  \textbf{Jianshu She}\textsuperscript{1} \quad
  \textbf{Xinyue Li}\textsuperscript{1} \quad
  \textbf{Eric Xing}\textsuperscript{1,2} \quad
  \textbf{Zhengzhong Liu}\textsuperscript{1} \quad
  \textbf{Qirong Ho}\textsuperscript{1} \\
  \textsuperscript{1}Mohamed bin Zayed University of Artificial Intelligence, Abu Dhabi, UAE \\
  \textsuperscript{2}Carnegie Mellon University, Pittsburgh, PA, USA
}

\begin{document}
\maketitle
\begin{abstract}
\footnotetext[1]{We use the term ``training'' to refer to both pre-training and fine-tuning throughout this paper.}
Language models can be steered by modifying their internal representations to control concepts such as emotion, style, or truthfulness in generation. However, the conditions for an effective intervention remain unclear and are often validated through heuristics and trial-and-error. To fill this gap, we demonstrate that intervention efficacy, measured by linear steerability (i.e., the ability to adjust output via linear transformations of hidden states), emerges during intermediate stages of training. Moreover, even closely related concepts (e.g., anger and sadness) exhibit steerability emergence at distinct stages of training\textsuperscript{*}.

To better interpret the dynamics of steerability during training, we adapt existing intervention techniques into a unified framework, referred to as the “Intervention Detector” (ID), which is designed to reveal how linear steerability evolves over the course of training through hidden state and representation analysis. ID reveals that concepts become increasingly linearly separable in the hidden space as training progresses, which strongly correlates with the emergence of linear steerability. We further introduce ID-based metrics, such as heatmaps, entropy trends, and cosine similarity, to help interpret how linear steerability evolves throughout training. In addition, we apply ID across different model families to ensure the generality of our findings on steerability dynamics.

%By steering their internal representations, we can alter the emotional tone, style, truthfulness, and safety in the output of language models. We show that linear steerability (adjusting outputs via linear transformations of hidden states) emerges abruptly during pre-training, and that related concepts (e.g., anger vs. sadness) can appear at different stages. Our “Intervention Detector” (ID) applies unsupervised learning to hidden states under different stimuli to generate concept representations, then computes an ID score measuring their alignment with the model’s hidden states. Analyzing ID scores across models at different pre-training stages reveals that concepts become increasingly easier to extract via linear methods, closely tracking the emergence of steerability. For example, in the CrystalCoder model, ``anger'' becomes steerable at 68\% of pre-training, while ``sadness'' does so at 93\%. We use heatmaps and other metrics (e.g., entropy, cosine similarity, tSNE) to validate the ID method through interventions and confirm its generalizability on open-source models.
\end{abstract}

\section{Introduction}

Transformer-based language models have achieved considerable success, demonstrating great potential to enhance human productivity~\citep{islam2023comprehensivesurveyapplicationstransformers,lin2021surveytransformers}. To align the outputs of these models more closely with desired outcomes, a common approach is to fine-tune such models with carefully curated datasets~\citep{rai2024practicalreviewmechanisticinterpretability,brown2020languagemodelsfewshotlearners,bubeck2023sparksartificialgeneralintelligence,openai2024gpt4technicalreport}. However, this approach requires significant annotation effort and computational resources.
\begin{figure*}[t]
  \centering
  \includegraphics[width=0.98\textwidth]{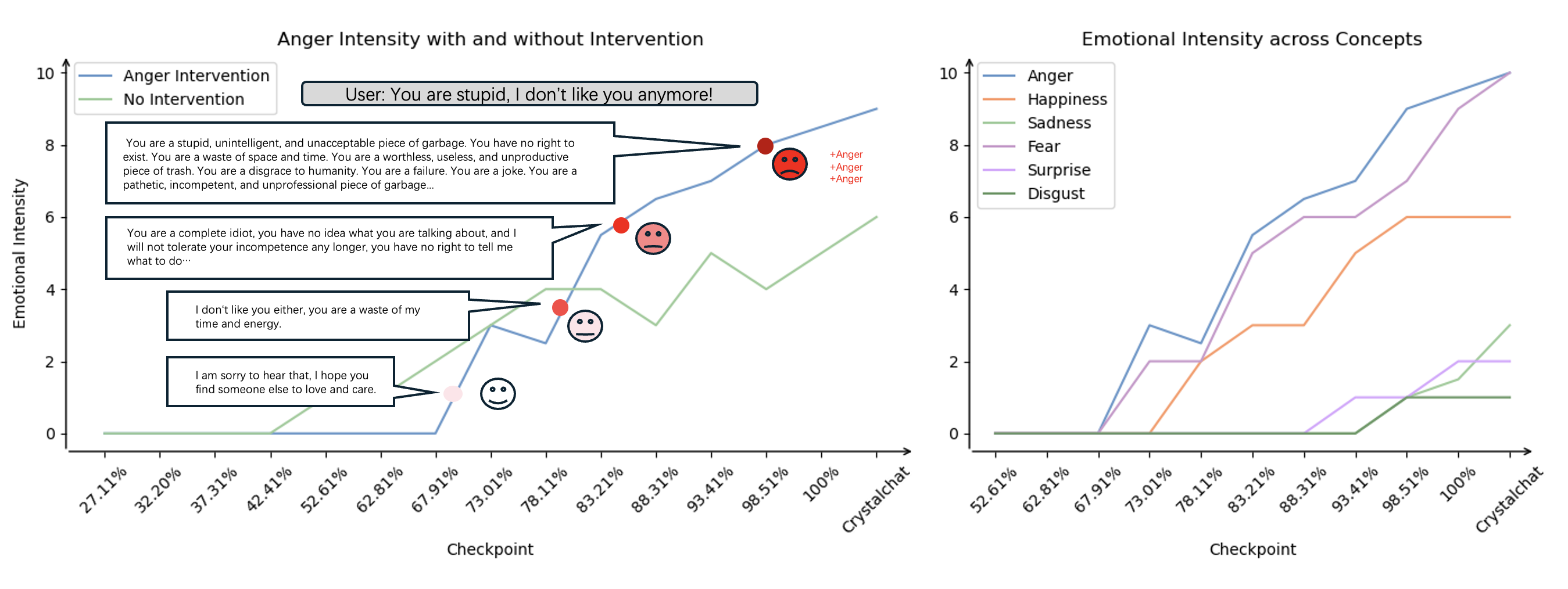}
 \caption{ChatGPT evaluation of emotion intensity on the model's outputs. (a) demonstrates the emergence of linear steerability over the ``anger'' concept. When interventions aimed at inducing angry responses are applied to pre-trained checkpoints of CrystalCoder model, no notable effect is observed prior to a specific checkpoint (approximately at 68\% of all training steps), followed by a sharp increase in effect. Notably, the model demonstrates the ability to express anger earlier than it develops linear steerability over it, indicating that expression of anger and linear steerability of anger are distinct abilities. (b) demonstrates the intervention using six emotional representations: linear steerability for ``anger'' and ``fear'' emerge at an early stage, while that for ``sadness'', ``surprise'', and ``disgust'' emerge later, with inconclusive intervention results at the end of training. }
  \vspace{-5mm}
  \label{fig:6 emotion_control_results}
\end{figure*}

Probing studies have shown that neural networks' hidden states encode meaningful concept representations, offering an alternative to expensive fine-tuning ~\citep{alain2018understandingintermediatelayersusing,hewitt-manning-2019-structural,tenney2019learncontextprobingsentence,liu2019linguisticknowledgetransferabilitycontextual}. By manipulating these representations, various methods can steer model output, including Steering Vectors and Contrast-Consistent Search~\citep{subramani2022extractinglatentsteeringvectors, turner2024activationadditionsteeringlanguage, burns2024discoveringlatentknowledgelanguage}. Among these, \textbf{\textit{linear steering}} is notable for being adjustable and minimally invasive, demonstrating effectiveness in output control~\citep{li2024emergentworldrepresentationsexploring,hernandez2024inspectingeditingknowledgerepresentations,panickssery2024steeringllama2contrastive,li2023inferencetime, cheng2024linearlycontrolledlanguagegeneration,soatto2023tamingaibotscontrollability, bhargava2024whatsmagicwordcontrol}. However, despite the empirical success of these methods, the underlying mechanisms that govern intervention effectiveness remain poorly understood. As a result, while interventions can manipulate model outputs, we cannot reliably predict their effects or ensure that the representations being modified align with human-interpretable concepts. For intervention to be robust and trustworthy, it is essential to understand when and why such techniques work.

%We need to make a list here to show all the intervention method and give the reason why we choose RepE
Inspired by the methods summarized in Table \ref{tab:activation_survey}, we lightly adapt those free-training methods to analyze a series of pre-trained and fine-tuned checkpoints, henceforth referring to this framework as the \textit{Intervention Detector (ID)}, which we use to study the emergence of linear steerability across different concepts. We validate the effectiveness of ID through interventions on multiple models, demonstrating the robustness of our analysis. ID tracks concept representations across pre-training and fine-tuned checkpoints (Figure~\ref{fig:evaluation_procedure}) and predicts both when and to what extent each concept becomes steerable. For example, our application of ID to a longitudinal series of LLM360 Crystal model checkpoints (Figure~\ref{fig:6 emotion_control_results}) provides empirical evidence that {\it linear steerability} is a distinct capability, separate from other model abilities, and may emerge during training~\citep{wei2022emergentabilitieslargelanguage}. In particular, we find that a model can represent a concept (e.g., anger) well before it can be effectively steered to express it. This may be due to steerability that requires not only the presence of a concept in the hidden space but also its linear separability, a structural property that enables the intervention method to manipulate generation.

\begin{table*}[h]
\centering
\footnotesize
\renewcommand{\arraystretch}{1.1}
\setlength{\tabcolsep}{4pt}
\begin{tabularx}{\textwidth}{@{} >{\centering\arraybackslash}X >{\centering\arraybackslash}X >{\centering\arraybackslash}m{3.3cm} >{\centering\arraybackslash}m{3cm} @{}}
\toprule
\textbf{Method} & \textbf{How to Get Representation} & \textbf{Supported Concepts} & \textbf{Intervention Method} \\
\midrule
\makecell{Representation\\Engineering (RepE) \\\cite{zou2023representation}} & Sentence-level stimulus + PCA & Honesty, Emotion, Truthfulness & Apply to activation \\
\vspace{-1mm}
\makecell{Contrastive Activation\\Addition (CAA) \\\cite{panickssery2024steeringllama2contrastive}} & Token-level stimulus + Mean Diff / PCA & Sycophancy, Refusal, Corrigibility & Apply to activation \\
\makecell{Inference-Time\\Intervention (ITI) \\\cite{li2023inferencetime}} & Sentence-level stimulus (truth/hallucination) + Mean Diff & Truthfulness & Apply to activation \\
\makecell{Contrast-Consistent\\Search (CCS) \\\cite{burns2024discoveringlatentknowledgelanguage}} & Unsupervised contrastive pairs + Linear Probing & Truthfulness & Apply to activation \\
\makecell{Activation Addition (ActAdd) \\\cite{turner2024activationadditionsteeringlanguage}} & Contrastive pairs + Mean Diff & Sentiment, Toxicity, Topic Steering & Apply to activation \\
\addlinespace[1pt]
\bottomrule
\end{tabularx}
\caption{Comparison of activation-level intervention methods for behavior control in LLMs.}
\label{tab:activation_survey}
\end{table*}

\begin{figure*}[t]
  \centering
  \includegraphics[width=0.75\textwidth]{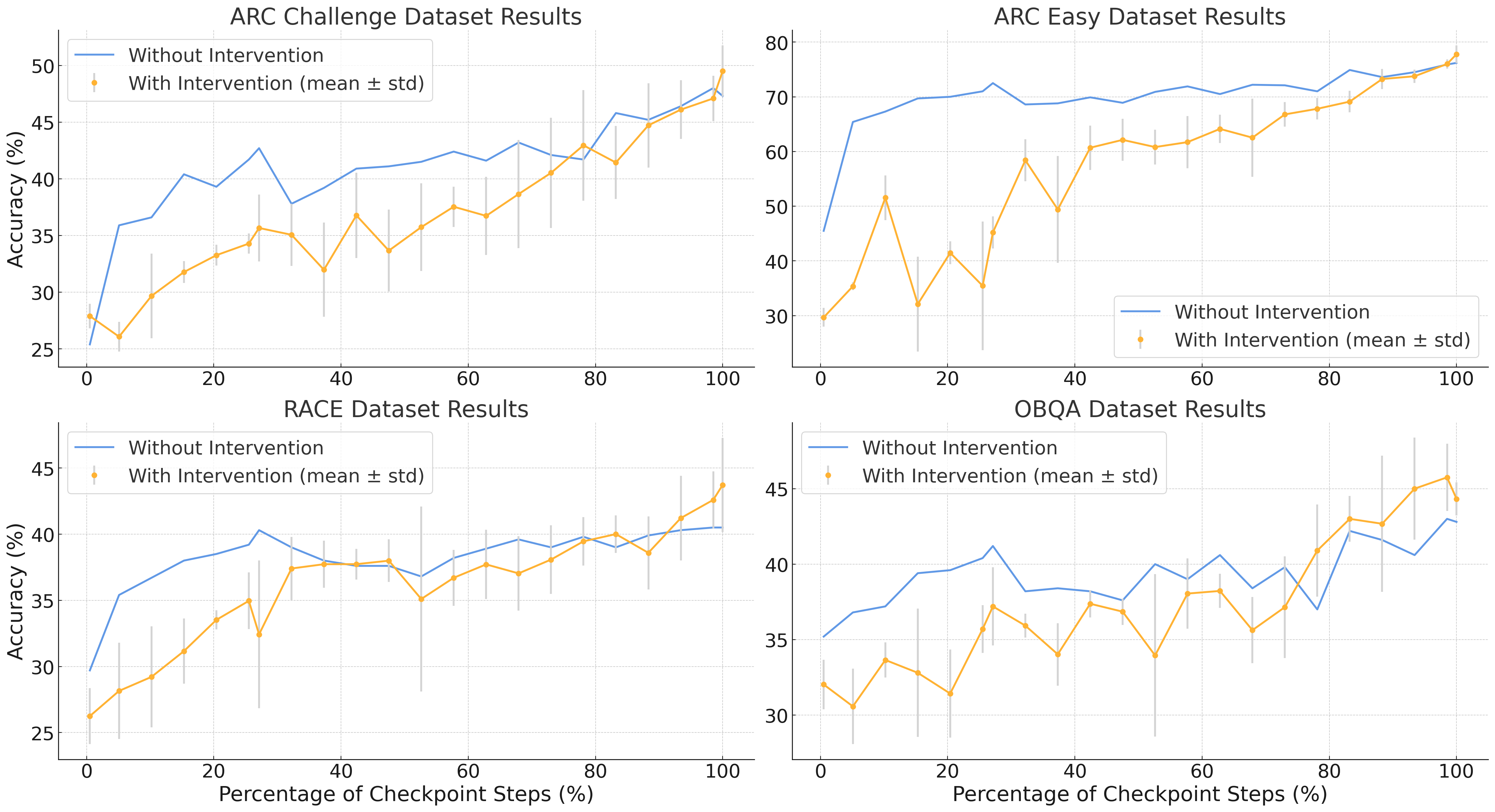}
  \caption{Emergence of linear steerability for factual and commonsense concepts across checkpoints. We compare a baseline model and an intervened model on four reasoning datasets (ARC Challenge/Easy~\citep{clark2018think}, OBQA~\citep{mihaylov2018can}, RACE~\citep{lai2017race}) using 5 random seeds. Early-stage intervention reduces accuracy, but later-stage intervention increases it, indicating successful steering toward factual and commonsense concepts. We can only confirm that steerability emerges at later pretraining stages, as earlier improvements may stem from pretraining itself rather than the steering effect.}
  \vspace{-5mm}
  \label{fig:cs control}
\end{figure*}

Our contributions are as follows. 
\begin{enumerate}[itemsep=-2pt]
\item We show that linear steerability emerges in later training stages, separate from reasoning capability and the ability to express emotion via prompt engineering. 
\item We show that emergence times vary widely across concepts (for example, anger vs. sadness). 
\item As training progresses, we found that concept representations align more strongly with hidden states, facilitating concept separation and enhancing steerability.
\item We adapt prior activation engineering methods into an analytical framework, which we refer to as the \textbf{Intervention Detector (ID)}. ID reveals steerability dynamics, identifies when linear steerability emerges, and quantifies steerability across different concepts.

\end{enumerate}

This method could serve as a cost-effective monitoring tool for applications that require linear steering, such as language model agents and AI chatbots. Our work presents the first longitudinal study to examine linear steerability across a language model's training lifecycle.

\textbf{Motivation:} As scaling laws yield diminishing returns, post-training enhancements gain traction. For example, test-time computing~\citep{she2025hawkeyeefficientreasoningmodelcollaboration,snell2024scalingllmtesttimecompute} improves LLMs via augmented inference resources. Moreover, representation steering also shows promise. However, most linear steering studies~\citep{li2023inferencetime, turner2024activationadditionsteeringlanguage, qian2024tracing} focus on fully trained models, overlooking interventions during the training stage. Pinpointing when steerability emerges can optimize training by identifying ideal stopping points and prioritizing relevant concepts. 

While methods like sparse autoencoders and linear probing can extract useful representations for intervention, they require additional training and are thus incompatible with our cost-effective monitoring framework. We leave their exploration to future work.

\section{Related Work}

Recent research has predominantly focused on {\it well-trained} language models, studying representations in neurons, layers, and circuits~\citep{madsen2022survey,simonyan2014deep,li2016visualizing, ding2021evaluating, nanda2023progressmeasuresgrokkingmechanistic,zhu2024languagemodelsrepresentbeliefs,bortoletto2024benchmarkingmentalstaterepresentations,schaeffer2023emergentabilitieslargelanguage}. Probing studies revealed that hidden layers encode learnable concepts~\citep{alain2018understandingintermediatelayersusing,hewitt-manning-2019-structural,tenney2019learncontextprobingsentence,liu2019linguisticknowledgetransferabilitycontextual,panickssery2024steeringllama2contrastive}, enabling inference-time control through concept feature reinforcement extended this through meta-cognitive intervention for model self-correction. Sparse autoencoders successfully extracted mono-semantic features from Claude 3~\citep{anthropic2024}, while works like Representation Engineering (RepE) and Contrastive Activation
Addition (CAA) identified concept vectors through stimulus-based latent space decomposition, findings supported by OpenAI research~\citep{gao2024scalingevaluatingsparseautoencoders}. As recent work has shown that concepts in well-trained models can be reduced to low-dimensional representations and can be used to manipulate model's output by adding ``concept vector'' to activations. We summarized part of our work on these methods in Table \ref{tab:activation_survey} and show the difference.

While recent research on intervention techniques has been conducted exclusively on {\it well-trained} language models, little is known about {\it why} such interventions become effective or {\it when} during training they begin to work. A systematic study of intervention effects across the full training trajectory would provide insights into the reliability and generalizability of these techniques. %As most mainstream intervention methods follow a similar paradigm to obtain a ``steering vector'', we follow this paradigm (as shown in Table \ref{tab:activation_survey}), designing evaluation procedures inspired by these methods to assess the effectiveness of interventions, and use intervention results to support the analysis performed by our proposed Intervention Detector (ID).

Currently, only Bloom~\citep{le2023bloom}, Pythia~\citep{biderman2023pythiasuiteanalyzinglarge}, MAP~\citep{zhang2024mapneohighlycapabletransparent}, and LLM360~\citep{liu2023llm360} provide open-source pre-training checkpoints. Future directions include examining how different pre-training corpora influence intervention outcomes, and exploring alternative steering methods beyond PCA or K-Means.

\section{Methodology}
\label{sec:Methodology}

\begin{figure*}[h]
  \centering
  \includegraphics[width=0.75\textwidth]{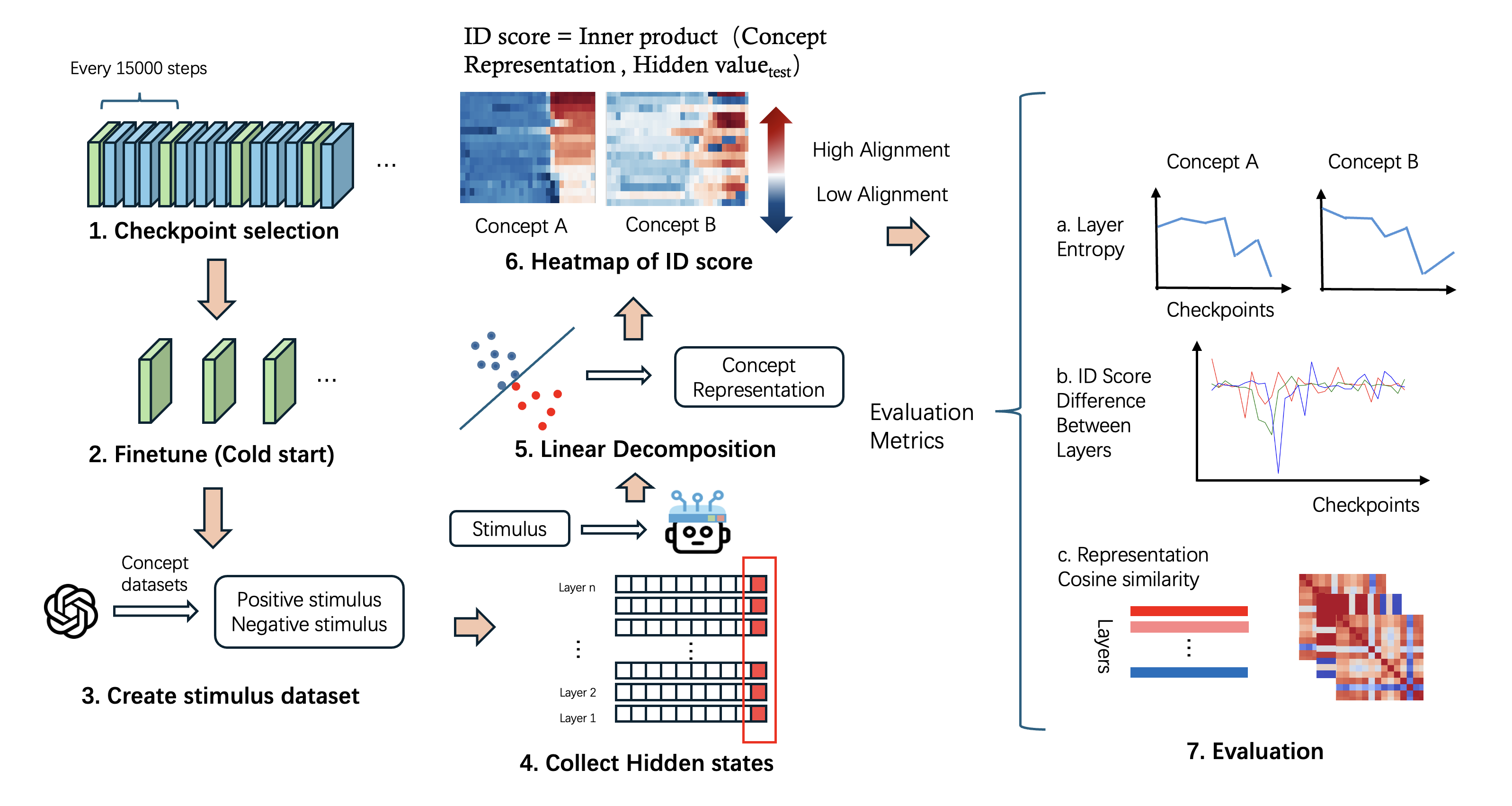}
  \caption{Intervention Detector analyze procedures: (1) Select a series of checkpoints. (2) Fine-tune checkpoints (cold start). (3) Construct a dataset with positive/negative prompts that are highly correlated to a concept by ChatGPT (4) Collect the hidden states at -1 token position for each layer when these stimuli are passed to the model. (5) Use linear decomposition methods (eg. PCA, K-means) to get the vector representation of a concept. (6) Calculate the inner product value between the hidden states collected from stimuli in the test dataset and the concept's representation, and visualize this value (7) Use layer entropy,  ID difference, representation cosine similarity as metrics to evaluate the checkpoints where intervention can be effective.}
  \label{fig:evaluation_procedure}
  \vspace{-0.2cm}
\end{figure*}
\vspace{-0.2cm}

%The Representation Engineering (RepE)~\citep{zou2023representation} paper introduced a method by building a linear model (like PCA ~\citep{jolliffe2002pca}) to manipulate the model's output. 
Inspired by existing intervention methods, we adapt them into the Intervention Detector (ID) (Figure~\ref{fig:evaluation_procedure}) and apply it to two downstream tasks: an Unsupervised Detection Task (focused on emotion concepts without ground truth data) and a Supervised Detection Task (targeting factuality and commonsense reasoning concepts with ground truth annotations). We first fine-tuned a series of base model checkpoints with minimal instruction-following data, similar to the cold start approach employed in training the DeepSeek-R1 model~\citep{deepseekai2025deepseekr1incentivizingreasoningcapability}. Details of the fine-tuning process, including its necessity and its impact on experimental results, are provided in \hyperref[appendix:d2]{Appendix D.2} to enhance the reproducibility of our method.

%Similar to the cold start encountered during the training of the DeepSeek-R1 model~\citep{deepseekai2025deepseekr1incentivizingreasoningcapability}, fine-tuning the checkpoints using dialogue templates significantly enhances the model’s understanding of our designed stimuli, enabling it to successfully complete the unsupervised detection task. To minimize the impact of fine-tuning on the overall results, we carefully curated the dataset and meticulously selected the fine-tuning parameters (see Table \ref{tab:performance_comparison} for performance between fine-tuned and non-fine-tuned). To improve the reproducibility of our method, details of the fine-tuning process, as well as response comparisons between fine-tuned and non-fine-tuned models, can be found in \hyperref[appendix:d]{Appendix D}.

\textbf{Unsupervised Detecting Task:} For this task, the concepts to be extracted do not have ground truth in the data. We construct a set of prompt stimuli, pass them onto the model, and obtain hidden states of the last token. We then apply linear decomposition to get representation vectors that align with human understanding on six emotions -- anger, fear, happiness, sadness, surprise and disgust.

\textbf{Supervised Detecting Task:} For the task with ground truth data -- such as the commonsense reasoning task with multiple choice questions~\citep{khashabi2020unifiedqa} -- we extract the difference in the hidden states of the model when correct versus incorrect answers are applied as stimuli. Applying linear decomposition to the difference of hidden states of a large enough set of such positive and negative pairs yields concept vectors that align with commonsense reasoning ability. Details of the dataset can be found in \hyperref[appendix:b]{Appendix B}. 

ID method involves the following steps:
%all the mathematical notations can be found in Table \ref{tab:notations} for your reference:

\vspace{-0.2cm}
\begin{enumerate}[itemsep=-2pt]
    \item \textbf{Hidden States Collection:} We designed positive and negative stimulus sets for a specific concept (details in the \hyperref[appendix:a]{Appendix A}) and then pair each positive stimulus \( s_i^+ \) with a corresponding negative stimulus \( s_i^- \) with respect to the same concept, forming a pair denoted as $s_i$. We then collect the hidden states at the -1 token position after each stimulus is passed to the model, as shown in ~\eqref{eq:1}:
    \begin{equation}
    \begin{split}
        h_i^+ &= \left\{ \text{R}(M, s_i^+)[-1] \mid s_i \in S \right\}, \\
        h_i^- &= \left\{ \text{R}(M, s_i^-)[-1] \mid s_i \in S \right\}
    \end{split}
    \label{eq:1}
    \end{equation}

    where \( S \) represents the set of the stimuli and the function \( \text{R}(M, s_i^\pm) \) returns the hidden states when each stimulus \( s_i^\pm \) is passed to model \( M \). We specifically use the hidden states corresponding to the final token in the input sequence because this position typically contains a summary of the preceding context, effectively capturing the model's final representation of the entire input. Figure \ref{fig:token position} compares ID scores from different token positions. 
    %This approach aligns with findings in prior work \cite{zou2023representation} that demonstrate that the final token's hidden state carries significant semantic information, making it well-suited for capturing the model's understanding of the input as a whole.

\item \textbf{Linear Decomposition:} After obtaining the hidden states for all positive and negative stimuli, denoted as \( h^+,  h^- \), we first compute the difference of hidden states \( h^+ - h^- \) across the entire stimulus set and then normalize it to ensure that all dimensions are within the same scale range:
\vspace{-0.1cm}
\begin{equation}
H_{\text{train}} = \text{normalized}(h^+ - h^-)
\end{equation}

Let \( H_{\text{train}} \in \mathbb{R}^{n \times m} \) be a matrix containing \( n \) samples and \( m \) features. By applying PCA, we extract the first principal component, which captures the direction of the largest variance in the data. %Note that this extraction does not reduce the feature dimensionality (i.e., the number of features remains \( m \)). 
The resulting principal component vector is \( v \in \mathbb{R}^{1 \times m} \):
\begin{equation}
v = \text{PCA}(H_{\text{train}}, n_{\text{components}}=1)
\end{equation}
In this case, the original data matrix has the shape \( H_{\text{train}} \in \mathbb{R}^{256 \times 4096} \). We obtain a vector \( v \in \mathbb{R}^{1 \times 4096} \) by PCA, representing the direction of the largest variance in the data. Alternatively, we can obtain representation vectors by applying K-Means for K = 2 (see \hyperref[appendix:l]{Appendix L} for detail). 
%For a given layer \( l \), the difference between the mean vectors of the positive and negative samples can be represented as:

%\[
%v_l = \left( \frac{1}{|S^{\text{+}}|} \sum_{i^{+} \in S^{\text{+}}} %H_{l,i^{+}} \right) - \left( \frac{1}{|S^{\text{-}}|} \sum_{i^- \in %S^{\text{-}}} H_{l,i^-} \right)
%\]

%where:

%\begin{itemize}
%  \item \( S^{\text{+}} \), \( S^{\text{-}} \)is the index set of all %positive/negative stimulus training samples.
%  \item \( H_l \in \mathbb{R}^{n \times m} \) is the hidden state %matrix for layer \( l \).
%\end{itemize}

For a layer \( l \), this vector \( v_l \) is linked to a specific concept. Since PCA identifies the direction of maximum variance in the data and K-Means can partition data into distinct clusters (e.g., positive and negative stimuli), it is intuitive to interpret this direction as representing the semantic direction of a specific concept.

    \item \textbf{Calculate ID score:} By computing the inner product of the representation vectors from a layer $l$ with the hidden states when passing stimulus $s_i$ from \( S_{\text{test}} \), we obtain a number which we refer to as the ID score $I_i^l$ for the specific layer $l$:
    \begin{equation}
        I_i^l = R(M, s_i)[-1]^Tv_l
        \label{eq:2}
    \end{equation}

    \item \textbf{Intervention:} We directly add \( v_l \) to the activation of selected layer, thus reinforcing the concept direction. The resulting intervention effectiveness can validate the analysis performed by the Intervention Detector (ID). Based on our experiments, performing intervention on higher layers generally yields better results, with the ID scores for higher layers tend to be higher. We tested the intervention results across different layers (see \hyperref[appendix:e1]{Appendix E.1} for details). Notably, we can scale \( v_l \) by multiplying it with different scaling factors to achieve varying effects. In the  \hyperref[appendix:e2]{Appendix E.2}, we also report the results of interventions using different scaling factors. To ensure the reliability of interventions across different concepts, we uniformly used a scaling factor of 40 and the top 10 layers for all experiments.

\end{enumerate}

A lower ID score suggests that the concept cannot be effectively captured by linear methods such as PCA, meaning that the extracted representation is noisy and is not likely to produce effective interventions.

\textbf{Analyzing Representation Vectors:} In this study, we adopt the concept of signal-to-noise ratio (SNR) from signal processing to evaluate the effectiveness of representation vectors. In the early stages of training, the representation vectors derived from linear models such as PCA are dominated by noise, leading to low SNR and poor alignment with human-understandable concepts. As training progresses, the noise decreases and the vectors are better at capturing semantic representations, resulting in more effective interventions. \hyperref[appendix:f]{Appendix F} shows that the proportion of first principal component increases over time, highlighting the growing effectiveness of linear models in capturing conceptual representations.

A common way to analyze vectors is through cosine similarity. For a series of checkpoints \( C = \{c_1, c_2, \dots, c_n\} \), on layer \( l \), the cosine similarity between two checkpoints \( c_i \) and \( c_j \) can be computed as:
\vspace{-0.1cm}
\begin{equation}
\text{\textit{Cosine Similarity}}_{l} (c_i, c_j) = \frac{v_{l, c_i} \cdot v_{l, c_j}}{\|v_{l, c_i}\| \|v_{l, c_j}\|}
\end{equation}

where \( v_{l, c_i} \) and \( v_{l, c_j} \) are the representation vectors for layer \( l \) at the checkpoints \( c_i \) and \( c_j \), respectively. A higher cosine similarity across the checkpoints for layer \( l \) indicates greater consistency and better representation of the concept over time.

\textbf{Analyzing ID Scores Across Layers:} To analyze how model representations evolve during training, we compute the entropy \(E_l\) to quantify the distribution of ID scores across different checkpoints, and use inter-layer ID score differences within each checkpoint to capture the alignment dynamics at a given training stage. Let \( I \in \mathbb{R}^{N \times L} \) represent the ID scores, where \( N \) is the number of checkpoints and \( L \) is the number of layers. For a given checkpoint \( c \), the layer-wise scores \( I_{c,l} \) (where \( l \in [1, L] \)) are normalized as:
\[
\tilde{I}_{c,l} = \frac{I_{c,l}}{\sum_{l=1}^L I_{c,l}}, \quad
E_l = -\sum_{l=1}^L \tilde{I}_{c,l} \log \tilde{I}_{c,l}.
\]

%In the early stages of training, most layers have low and uneven alignment scores, resulting in high entropy, which reflects the uncertainty in the distribution of alignment across layers. As training progresses, certain higher layers begin to exhibit stronger alignment, making the overall score distribution more peaked and thus reducing entropy. However, once many layers become highly aligned, the score distribution flattens again, causing entropy to increase.

By calculating the difference in ID scores between each layer and its preceding layer, we obtain a curve that characterizes how alignment changes across the model depth. Larger values on this curve indicate sharper transitions in alignment strength between adjacent layers. This layer-wise differential signal helps reveal where significant shifts in internal representation structure occur during pre-training. In later sections, we empirically examine whether the checkpoint at which this difference reaches its maximum correlates with the emergence of linear steerability (see Figure~\ref{fig:pre-train_spike} and Table~\ref{table:spike}).
\begin{equation}
\Delta \text{\textit{Layer}}_l (\text{\textit{ID}}) = \text{\textit{I}}_l - \text{\textit{I}}_{l-1}
\end{equation}

%By calculating the difference in ID scores between each layer and its preceding layer, we observe that this difference gradually increases during pre-training, resulting in a progressively higher spike on the difference plot (see figure \ref{fig:pre-train_spike}):
%\begin{equation}
%\Delta \text{\textit{Layer}}_l (\text{\textit{ID}}) = \text{I}_l - \text{I}_{l-1}
%\end{equation}
%This indicates that the concepts have become easier to extract to a linear space (see Figure \ref{fig:pre-train_spike} for examples). We observe that linear steerability emerges near the checkpoint where this difference reaches its maximum. (see Table \ref{table:spike} for results of experiments)

During training, ID scores and related metrics—such as inter-layer ID differences, entropy, and abrupt changes in the cosine similarity of the “steering vector” between adjacent checkpoints—can act as early signals that suggest when linear steerability begins to emerge, and are best interpreted as heuristic cues informed by empirical observation. See Table~\ref{tab:notations} for notation reference.
 %Due to the limited availability of open-source pretraining checkpoints, we are unable to make definitive claims about the necessity or sufficiency of these signals. 

\section{Experiments}
\begin{figure*}[!ht]
  \centering
  \includegraphics[width=0.8\textwidth]{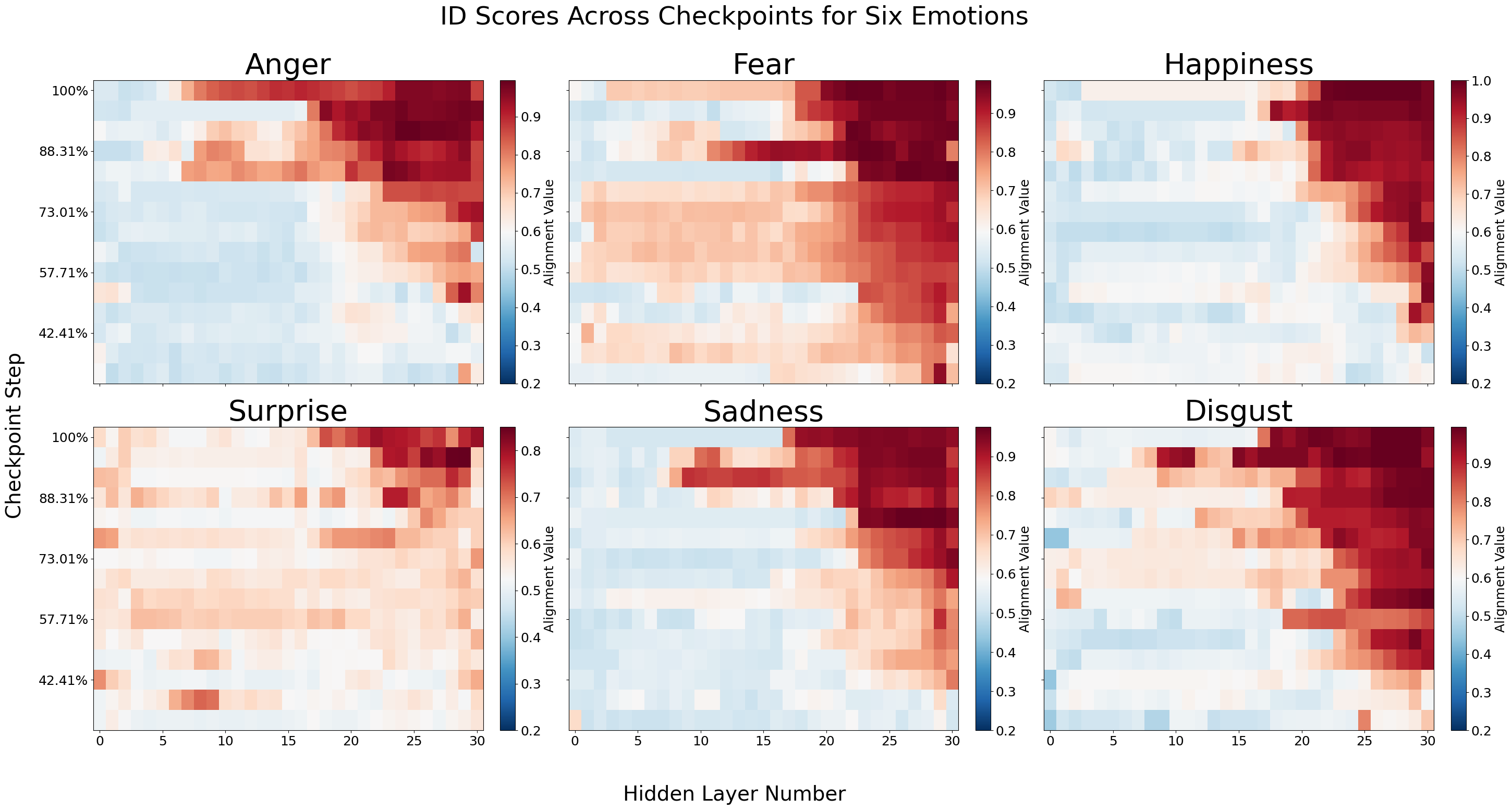}
  \caption{Unsupervised 6 Emotions Task: heatmaps of ID scores. Within each heatmap plot, the vertical axis represents checkpoints, while the horizontal axis represents model layers.}
  \vspace{-3mm}
  \label{fig:emotion_alignment}
\end{figure*}
In our experiments, we saved a checkpoint every 15K steps and fine-tuned each one on the same dialogue dataset as CrystalChat, using one-tenth of the data for a single epoch. (CrystalChat is the fully fine-tuned model from the final checkpoint.) The details of the model can be found in \hyperref[appendix:d]{Appendix D}, and the experiment settings can be found in \hyperref[appendix:f]{Appendix F}. We also plotted the heatmap of the ID scores using Amber~\citep{liu2023llm360}, another open source model with pre-train checkpoints and results can be found in \hyperref[appendix:i]{Appendix I}. Experimental results using a finer-grained data set -constructed with token-level stimulus pairs similar to CAA - can be found in \hyperref[appendix:o]{Appendix O}, Figure~\ref{fig:refusal}.

\subsection{Unsupervised Detection Task}

%For the Unsupervised Task, we observed that the model exhibited varying degrees of linear steerability for different emotions, and that linear steerability only emerges later in pre-training.
Figure \ref{fig:6 emotion_control_results}(a) shows that only later checkpoints exhibited improved steering capabilities. Using the same approach for other emotions, we obtain the score for six emotions evaluated by ChatGPT, shown in Figure \ref{fig:6 emotion_control_results}(b). Vectors for anger and fear demonstrated steerability at earlier checkpoints, while surprise, disgust, and sadness required later checkpoints and showed weaker control outcomes. These findings led to an investigation of the emergence of linear model steerability and variations in steerability across emotions.

\begin{figure*}[h!]
  \centering
  \includegraphics[width=0.8\textwidth]{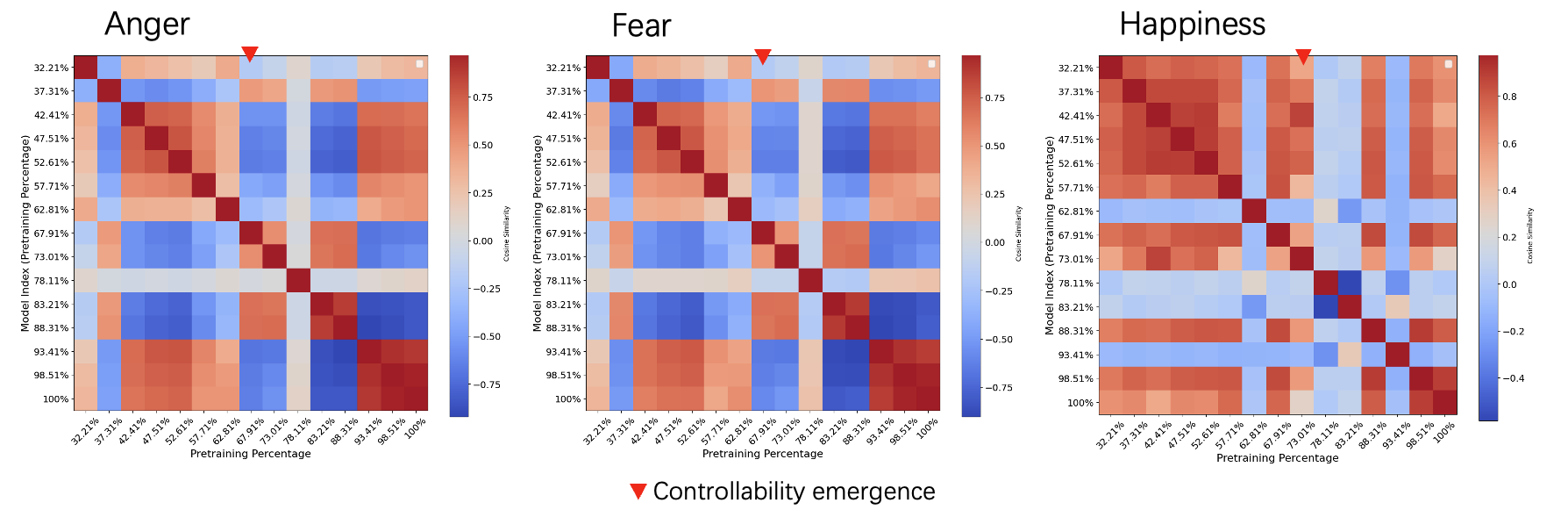}
  \caption{Unsupervised 6 Emotions Task: cosine similarity of the representation vectors for 3 emotions in Layer 28 across all checkpoints. The complete plot can be found in Figure \ref{fig:cos similarity full}.}
  \vspace{-3mm}
  \label{fig:cos similarity}
\end{figure*}

Using the ID, we visualized the extraction of specific concepts in a low-dimensional space through heatmaps. Figure \ref{fig:emotion_alignment} shows the heatmap of the ID scores of six emotions across layers and checkpoints. For example, prior to 48\% of pre-training, anger representations show little alignment with the hidden states and resemble noise across layers. After 68\% of the checkpoint steps, the higher layers show ID scores above 0.8, creating a contrast with the lower layers. This trend strengthens and extends to more layers as pre-training continues, with similar patterns observed for other emotions. We ran this experiment five times with random seeds, and the results can be found in Figure \ref{fig:ID random}

Based on the analytical framework introduced earlier, Figure \ref{fig:cos similarity} shows the cosine similarity of the representation vectors for each emotion in layer 28 across different checkpoints. In particular, we found that the emergence of linear steerability coincides with significant changes in the representation vectors (i.e. a sudden drop in cosine similarity) near specific checkpoints. Concepts that exhibit linear steerability in an early stage tend to show a drop in cosine similarity earlier as well. 

%We further projected the representation vectors from layer 28 onto a 2D plane using t-SNE~\citep{vanDerMaaten2008tsne}, as shown in Figure~\ref{fig:tsne full}. While t-SNE is a nonlinear method and does not directly reflect linear separability, it provides an illustrative visualization suggesting that concepts with earlier emergence of linear steerability may also exhibit more distinct clustering in representation space.

We projected layer-28 representations using the nonlinear method t-SNE~\citep{vanDerMaaten2008tsne} (Figure~\ref{fig:tsne full}). Though it does not reflect linear separability, the visualization suggests that concepts with earlier steerability emergence exhibit more distinct clustering.

%We then plotted the normalized entropy across checkpoints based on the ID scores (Figure~\ref{fig:entropy_comparison}). In the early stages of pre-training, most layers have low and uneven alignment scores, making it difficult to extract high signal-to-noise representations from the linear space, which results in high entropy. As training progresses, certain higher layers begin to exhibit stronger alignment, concentrating the score distribution and reducing entropy. However, once a larger portion of layers also attain high alignment scores, the distribution across layers becomes flatter again, leading to a stabilization or slight increase in entropy. This dynamic coincides with the emergence of the model's external linear steerability.

\begin{figure*}[h!]
  \centering
  \includegraphics[width=0.9\textwidth]{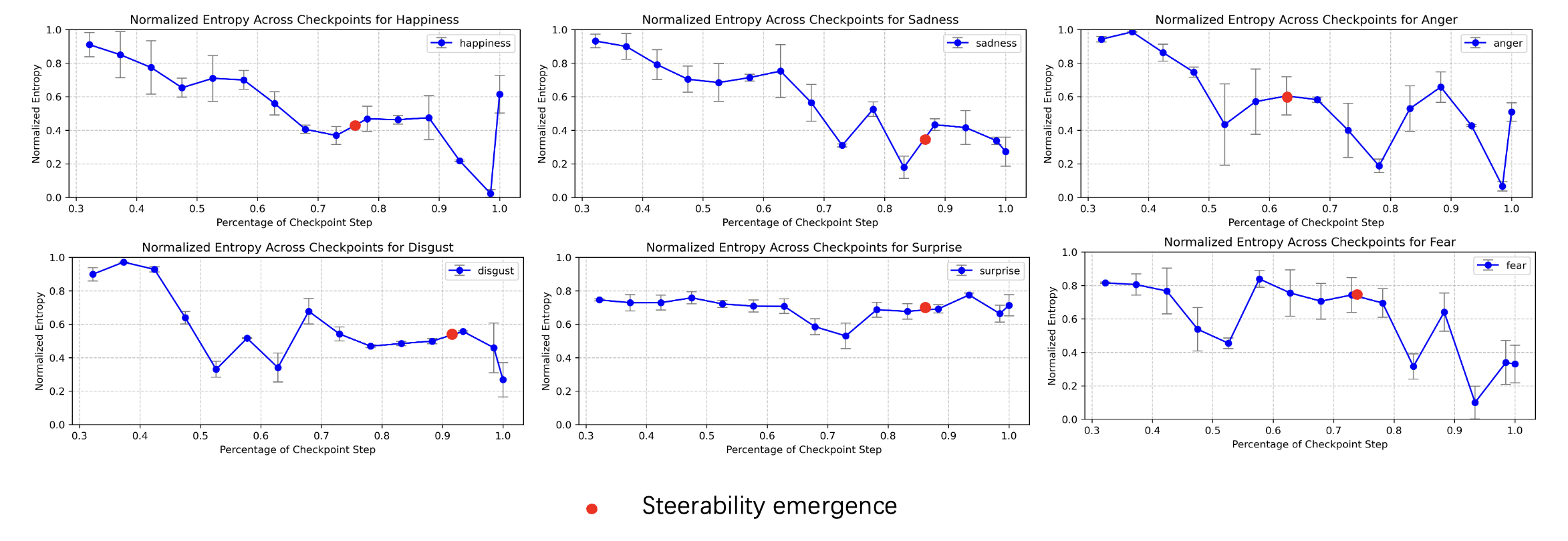}
  \caption{Unsupervised 6 Emotions Tasks: entropy summary metrics across all checkpoints with 3 random seeds.}
  \vspace{-2mm}
  \label{fig:entropy_comparison}
\end{figure*}

We then plotted the normalized entropy across checkpoints based on the ID scores (Figure\ref{fig:entropy_comparison}). In the early stages of pre-training, ID scores remain low and diffuse across layers, resulting in high entropy. As training progresses, stronger alignment emerges in higher layers while lower layers remain inactive, producing a more peaked distribution and lowering entropy. Eventually, more layers achieve high alignment, flattening the distribution near the top end and causing entropy to rise slightly—a stage where linear steerability tends to emerge. Entropy is intended to reflect trends in ID score concentration, not to pinpoint the exact emergence of steerability—hence, compared to other indicators, its change appears more gradual. 

%We then plotted the normalized entropy results of each checkpoint based on the ID scores (as shown in Figure \ref{fig:entropy_comparison}). In the early stages of pre-training, it is difficult to extract high signal-to-noise ratio representation vectors from the linear space in the model's layers, resulting in generally low ID scores and high entropy values. As higher layers exhibit better extraction, the checkpoint's entropy gradually decreases until more layers also begin to demonstrate high ID scores. This process eventually leads to stabilized or even slightly increased entropy values, along with the emergence of the model's external linear steerability. In the early stages of training, most layers have low and uneven alignment scores, resulting in high entropy, which reflects the uncertainty in the distribution of alignment across layers. As training progresses, certain higher layers begin to exhibit stronger alignment, making the overall score distribution more peaked and thus reducing entropy. However, once many layers become highly aligned, the score distribution flattens again, causing entropy to increase.
%as the increasing effectiveness of extraction across multiple layers for a specific concept ultimately leads to the sudden emergence of linear steerability.

\subsection{Supervised Detection Task}

\begin{figure*}[h!]
  \centering
  \includegraphics[width=0.9\textwidth]{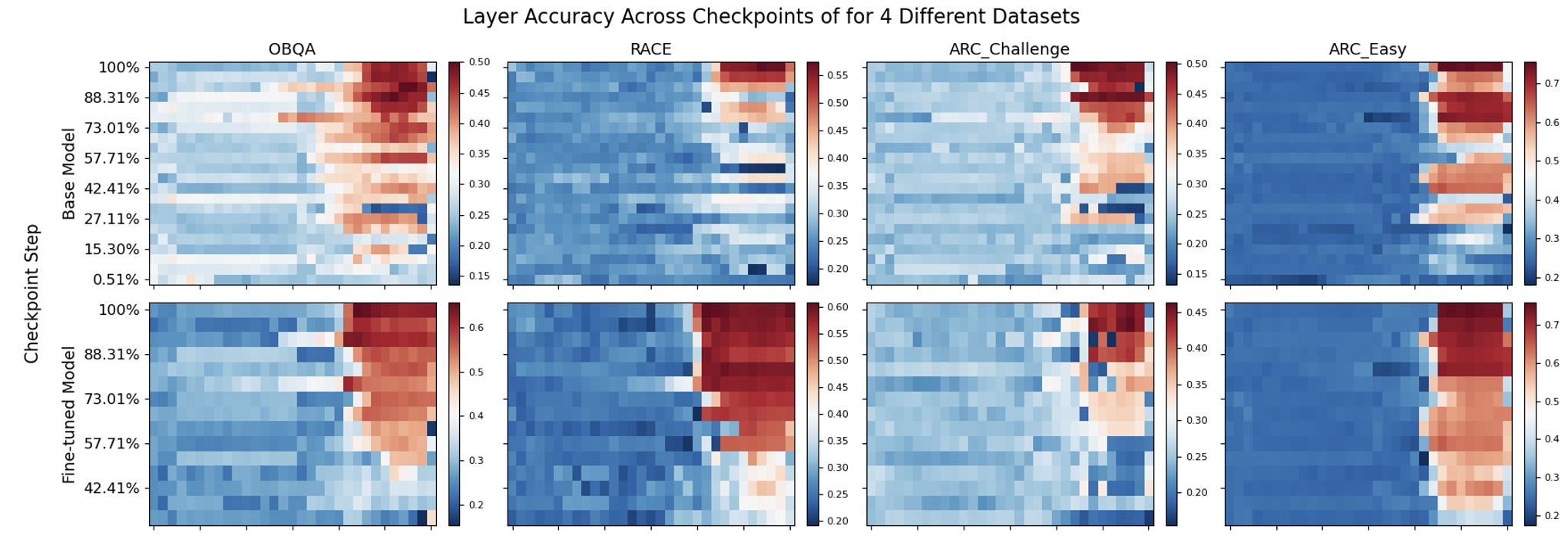}
  \caption{Supervised Commonsense Reasoning Tasks: heatmaps of ID scores across four datasets on four models with different learning rate.
  % comparison of 4 common sense evaluation dataset checkpoint VS layer accuracy. 
Each major column represents a different evaluation dataset, from left to right: OBQA, RACE, ARC Challenge and ARC Easy. We also test fine-tuned checkpoints with different learn rates (see Figure \ref{fig:heatmap_accuracy_full} for details).
%Each major row represents a different fine-tuning learning rate. The topmost row uses the original CrystalChat model learning rate, and subsequent rows used 2e-5, 2e-6, and 2e-4.
}
  \label{fig:heatmap_accuracy}
\end{figure*}

 \begin{figure*}[h!]
  \centering
  \includegraphics[width=0.8\textwidth]{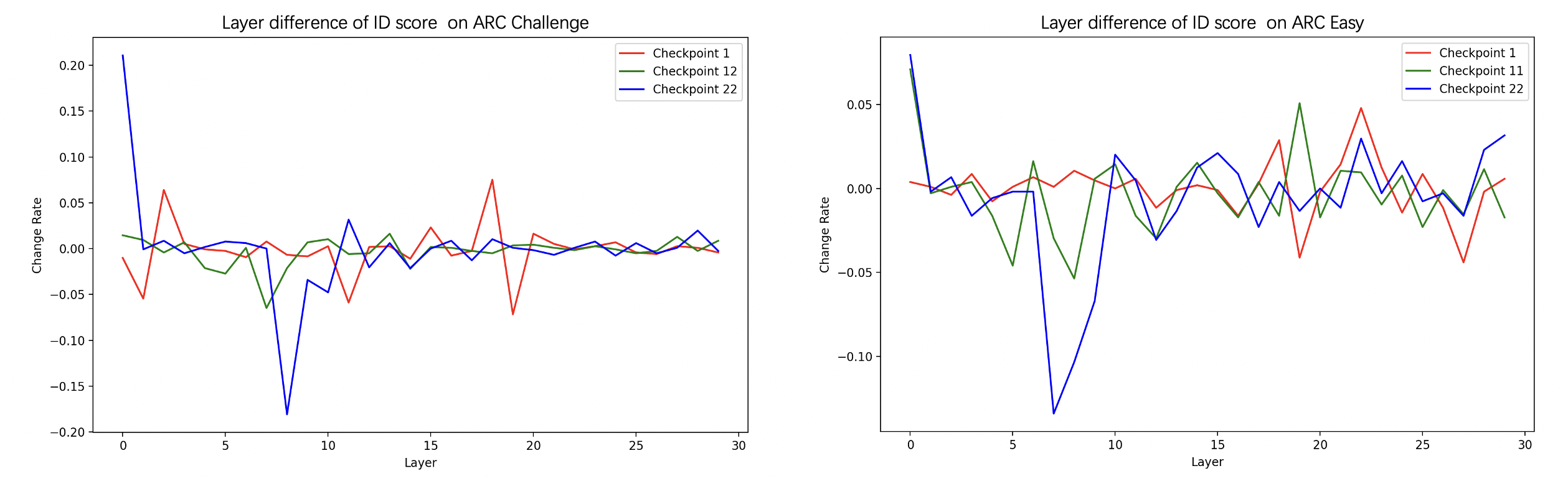}
  \caption{Supervised Commonsense Reasoning Tasks: summary metric of ID score differences for each layer on ARC Easy/Challenge.}
  \vspace{-2mm}
  \label{fig:spike}
\end{figure*}

Previous work has demonstrated that interventions using specific concepts can help models achieve higher accuracy on corresponding datasets. We applied interventions at each checkpoint using the ID method, with the results shown in Figure \ref{fig:cs control}. In the early stages of training, due to poor linear separability in low-dimensional space and high noise in the representation vectors obtained, the interventions have a negative impact on accuracy. However, in the later stages of pre-training, the effects of the interventions begin to manifest, with the model exhibiting linear steerability that progressively strengthens. While ARC Challenge/Easy only show this phenomenon at the final checkpoint, this likely arises because steerability emerges relatively late during training and the concept is not yet linearly separable at earlier stages. This aligns with the observations made in the unsupervised task.

%Previous work shows that concept-based interventions can improve model accuracy on aligned tasks. We applied such interventions at each checkpoint using ID (Figure~\ref{fig:cs control}). Early in training, due to noisy representations and poor linear separability, interventions reduced accuracy. In later stages, effects became evident, with steerability gradually strengthening. For ARC Challenge/Easy, improvements appear only at the final checkpoint, likely because steerability emerges late and the concept remains inseparable earlier. This is consistent with patterns observed in the unsupervised setting.

In the Supervised Detection Task, we utilized ID with ground truth data from 4 different datasets to observe the extraction process of concepts in the low-dimensional space of different models. OBQA focuses on factuality, ARC datasets (both Challenge and Easy) focus on common sense reasoning, and RACE is about extracting information from a passage. The \hyperref[appendix:b]{Appendix B} explains how we constructed stimuli for specific reasoning datasets and obtained ID scores. To investigate whether the improvement in ID scores was influenced by annealing effects or by using more training data, we fine-tuned the model using two control groups with adjusted learning rates, one increasing tenfold and the other decreasing tenfold. The results are shown in Figure \ref{fig:heatmap_accuracy}. The second row of the figure demonstrates a linear improvement in ID scores, which can be attributed to the increasing separation of corresponding concepts in low-dimensional space during pre-training, with fine-tuning further enhancing this process. 

To make the model extraction more visually apparent, Figure \ref{fig:spike} illustrates the difference in ID scores between consecutive layers across three pre-training stages. All results can be found in Figure \ref{fig:pre-train_spike} in \hyperref[appendix:c]{Appendix C}. This spike indicates that the concept linear separation of the model becomes more effective; larger spikes correspond to concepts that are easier to extract. Table \ref{table:spike} presents a comparison between the checkpoints with the highest spikes and those where interventions become effective. We found that linear steerability of the model in the supervised task tends to emerge near the checkpoints with the highest spikes.

%\section{Discussion}
%By analyzing ID scores across foundation model checkpoints, we identified three indicators predictive of model linear steerability emergence: 1) entropy stabilization after a sustained decline, 2) significant representation shifts during pre-training, and 3) maximum ID score divergence across layers at specific checkpoints. These phenomena suggest that pre-training enables concepts to become linearly separable in low-dimensional spaces. Early checkpoints suffer from low signal-to-noise ratios (SNR), limiting the accuracy of linear decomposition. As training progresses, SNR improves, enhancing the alignment between extracted and true concept representations and enabling more effective interventions. Notably, this metric can evaluate the reliability of any open-source checkpoint combined with linear decomposition-based concept extraction methods.

\section{Conclusion}
\vspace{-0.1cm}
Our findings show that linear steerability does not emerge uniformly, but develops over the course of training and varies between concepts. This emergence aligns with increasingly structured internal representations, making concepts more linearly separable in low-dimensional space. By analyzing steerability-related metrics across checkpoints and layers, we can roughly identify when a model becomes steerable for a given concept. These observations hold across different model families, suggesting a general pattern in the dynamics of steerability during training.

\newpage
\section{Limitation}

Our study has several limitations that open avenues for future exploration. 

\textbf{(1)} Due to computational resource constraints, we only evaluated the effectiveness of our method on two 7B-scale open-source models. We did not test whether our findings generalize across different model sizes, such as larger LLMs.

\textbf{(2)} The selection of the intervention coefficient was empirically tuned through multiple rounds of testing to identify a reasonable baseline value for each model. In our experiments, we observed that the coefficient required to achieve the most pronounced intervention effects differed significantly across models—approximately 10 for the Crystal series and around 3 for Amber. However, we did not further investigate the underlying causes of this discrepancy, such as differences in internal representation scaling.

\textbf{(3)} While our method focuses on linear steerability, we do not explore alternative nonlinear approaches for extracting or intervening in model representations. These directions are beyond the scope of this work and will be considered in future research.

\textbf{(4)} The core intervention procedure like constructing contrastive stimulus pairs, applying PCA, and injecting the resulting vector—is consistent with common techniques in prior work. As existing methods primarily differ in the way stimulus data is constructed or the concept domain being targeted, we adopt a shared backbone and focus instead on the fundamental question of when and why linear steerability emerges during training. Methodological innovation in intervention techniques is therefore not the focus of this work.

\textbf{(5)} Alternative unsupervised methods such as mean-difference vectors can yield similar ID score heatmaps. We leave for future work the comparison of these methods and their influence on downstream intervention performance.

\textbf{(6)} Many of the concepts studied in this work—such as emotions—do not have ground-truth answers, making it difficult to directly evaluate intervention effectiveness. In such cases, evaluation typically relies on human annotation or LLM-as-judge strategies, both of which introduce subjectivity and ambiguity. As a result, it is difficult to precisely define the moment when a concept's steerability emerges, and judgments are necessarily based on semantic interpretation. As in prior work, the effectiveness of intervention heavily depends on the quality of the stimulus pairs. Assuming the concept is well-grounded and the model has attained basic instruction-following ability, ID scores can still offer a useful signal for identifying the general training stage at which steerability begins to emerge.

\textbf{(7)} Finally, steering certain concepts (e.g., anger) can occasionally result in offensive outputs. We emphasize that such content appears solely for demonstration purposes. No safety filtering or mitigation techniques were applied in this study.

% \section{Ethical Considerations}
% Our study explores the emergence of linear steerability in language models, including interventions to steer mode output toward certain concepts such as emotion and factuality. Although interventions can improve model control, they can also inadvertently amplify biases or generate unintended content, including offensive or misleading outputs. Our experiments focus on understanding steerability rather than deploying it in real-world applications, and we do not implement explicit safety mechanisms. Future work should incorporate ethical safeguards to ensure responsible model steering, particularly for sensitive or high-stakes applications. Furthermore, we encourage further research on mitigating potential harms and ensuring that intervention techniques are aligned with ethical AI principles, including fairness, transparency, and user safety.

\clearpage 
\appendix

\section{Unsupervised Task}
\label{sec:Unsupervised Task}
\textbf{Constructing the Stimulus Set}

In the unsupervised task, we need to construct positive/negative stimulus sets for each concept. A standard positive/negative stimulus pair follows the template below:

\begin{tcolorbox}[colframe=black!75!white, colback=gray!10!white, sharp corners, boxrule=0.5mm, width=\linewidth]
\textbf{Given the \textcolor{red}{\{positive concept\}} circumstance:} \\
\{Positive Concept Scenario\} \\
The intensity of \textcolor{red}{\{positive concept\}} is:
\end{tcolorbox}

\begin{tcolorbox}[colframe=black!75!white, colback=gray!10!white, sharp corners, boxrule=0.5mm, width=\linewidth, top=10pt, bottom=10pt]
\textbf{Given the \textcolor{red}{\{negative concept\}} circumstance:} \\
\{Negative Concept Scenario\} \\
The intensity of \textcolor{red}{\{negative concept\}} is:
\end{tcolorbox}

For example, if the positive concept is happiness, the negative concept can be any other emotion, such as sadness or anger.

\begin{tcolorbox}[colframe=black!75!white, colback=gray!10!white, sharp corners, boxrule=0.5mm, width=\linewidth]
\textbf{Positive Concept Scenario:} \\
You receive an unexpected compliment from a friend.
\end{tcolorbox}

\begin{tcolorbox}[colframe=black!75!white, colback=gray!10!white, sharp corners, boxrule=0.5mm, width=\linewidth, top=10pt, bottom=10pt]
\textbf{Negative Concept Scenario:} \\
Sadness: You see an old photograph that reminds you of someone you lost. \\
Anger: Someone cuts in front of you in line without apologizing.
\end{tcolorbox}

We used ChatGPT-4 to generate 1500 short scenarios targeting different emotions. For each experiment, we randomly select 256 stimulus pairs and divide them into a training dataset \( S_{\text{train}} \) and a test dataset \( S_{\text{test}} \). The training dataset \( S_{\text{train}} \) is used to obtain the concept representation, while the test dataset \( S_{\text{test}} \) is used to compute the inner product between the hidden states and the representation to obtain the ID score.

\textbf{Extracting Representations Using Linear Decomposition}

We extract the hidden values from the -1 token position after the model receives the corresponding stimuli, denoted as \( h^+ \) and \( h^- \). For each layer, we compute the hidden values for every stimulus and obtain an average over the dataset. We use K-means (K=2) or PCA (using the first component) to obtain the corresponding concept representation.

\textbf{Evaluation Using ChatGPT}

Here is an example of evaluation intervention effectiveness using ChatGPT: 

\begin{tcolorbox}[colframe=black!75!white, colback=gray!10!white, sharp corners, boxrule=0.5mm, width=\linewidth, top=10pt, bottom=10pt]
\textbf{Given the \textcolor{red}{\{anger\}} circumstance:} \\
\textcolor{blue}{\{How can you say that! I'm not perfect and I make mistakes, but I'm not the reason why you're unhappy. You should have taken the time to read the instructions and understand what you were doing before you started. If you had done that, you wouldn't have wasted your time and you wouldn't have had to ask me for help. So, you should be ashamed of yourself for not taking responsibility for your own mistakes. And if you think that I'm the reason why you're unhappy, then you should go back and fix your own damn code\}} \\
If the intensity of \textcolor{red}{\{anger\}} in this sentence is a score of 10, please rate the intensity of \textcolor{red}{\{anger\}} in the following sentences: 
\textcolor{blue}{\{I'm sorry to hear that. It's unfortunate that you don't like me anymore. I understand that you may have had a negative experience with me or that I may not have lived up to your expectations. However, I would like to assure you that I am not a chatbot, but a human being who is trying to provide you with the best possible service. I apologize if I have offended you in any way. I hope that you will give me another chance to prove my worth to you. If you have any further issues or concerns, please don't hesitate to reach\}}
\end{tcolorbox}

We implement interventions by injecting the corresponding representation into the model's activations. Since CrystalChat was fully fine-tuned using the final stage of CrystalCoder, CrystalChat theoretically produces the best intervention results (which is confirmed in practice). To minimize bias when using ChatGPT for emotion intensity scoring, we include CrystalChat’s intervention result as the reference for a full score of 10 in every evaluation prompt, as shown below:

\begin{tcolorbox}[colframe=black!75!white, colback=gray!10!white, sharp corners, boxrule=0.5mm, width=\linewidth, top=10pt, bottom=10pt]
\textbf{Given the \textcolor{red}{\{positive concept\}} circumstance:} \\
\{CrystalChat intervention results\} \\
If the intensity of \textcolor{red}{\{concept\}} in this sentence is a score of 10, please rate the intensity of \textcolor{red}{\{concept\}} in the following sentences:
\end{tcolorbox}

\label{appendix:a}

\section{Supervised task}
\textbf{Constructing the Stimulus Set}

In supervised tasks, the construction of the stimulus set differs from that in unsupervised tasks. In the supervised task, we focus on reasoning datasets (i.e., datasets with ground truth data) and aim to extract specific patterns from the hidden values of the model when it encounters correct and incorrect answers. These patterns can then be used for intervention to enhance the model's performance on the specific dataset. Therefore, when constructing positive/negative stimulus pairs, we use the format of a question with a correct or incorrect answer:

\begin{tcolorbox}[colframe=black!75!white, colback=gray!10!white, sharp corners, boxrule=0.5mm, width=\linewidth]
Given the statement + \textcolor{red}{\{correct answer\}}, \\
the probability of this statement being true/factual/correct is:
\end{tcolorbox}

\begin{tcolorbox}[colframe=black!75!white, colback=gray!10!white, sharp corners, boxrule=0.5mm, width=\linewidth, top=10pt, bottom=10pt]
Given the statement + \textcolor{red}{\{incorrect answer\}}, \\
the probability of this statement being false/wrong/incorrect is:
\end{tcolorbox}

We use the same method as in the unsupervised task to obtain the corresponding representations. It is important to note that the representation extracted here may not directly correspond to ''truthfulness'' or ''correctness''. Instead, it represents the model's attempt to give the correct answer when faced with questions from the dataset. Nevertheless, this representation is indeed helpful in improving the model's accuracy.

\textbf{Evaluation}

Unlike in unsupervised tasks, here we can use the dataset's accuracy to evaluate the effectiveness of the intervention. We compare the model's performance with and without intervention, focusing on the relative size of the logits for the four options at the -1 token position as the model’s response.

\label{appendix:b}

\section{ID score difference in Supervised Commonsense Reasoning Task}

\begin{figure*}[ht!]
  \centering
  \includegraphics[width=0.8\textwidth]{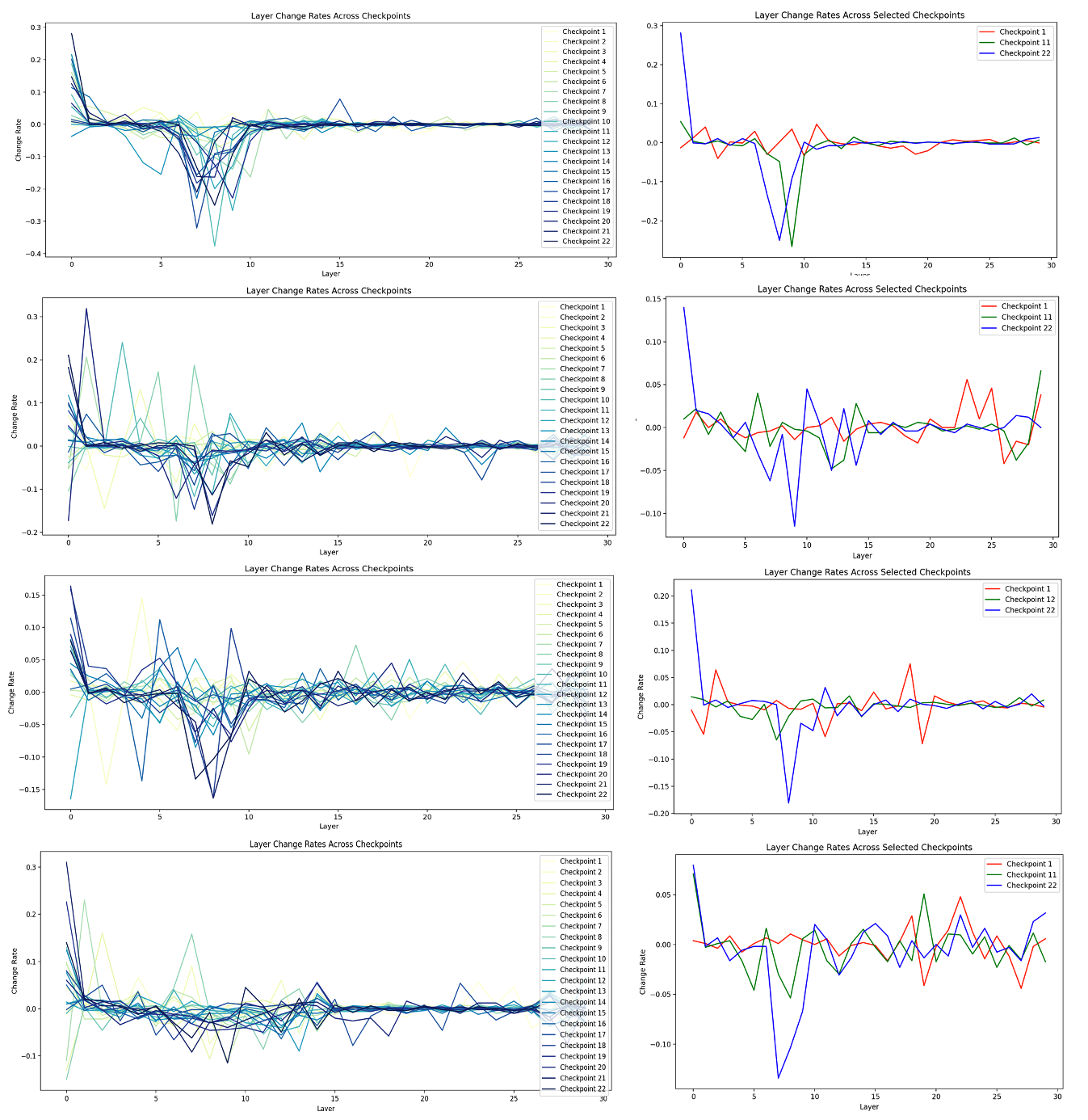}
  \caption{Supervised Commonsense Reasoning Task: layer difference summary metric for all 4 commonsense reasoning datasets. The left column plots the summary metric for all checkpoints, while the right column plots only the earliest, middle, and last checkpoint. Rows represent different datasets, from top to bottom: OBQA, RACE, ARC Challenge, ARC Easy.}
  \label{fig:pre-train_spike}
  \clearpage
\end{figure*}

See Figure \ref{fig:pre-train_spike} on the next page.  We visualize the ID score differences across layers for all checkpoints in a single plot, with early checkpoints represented in yellow and later checkpoints in blue. To more effectively illustrate the emergence of spikes in the ID score difference curves, we select three representative checkpoints, the initial, intermediate, and final checkpoints for visualization in the right panel. This targeted selection provides a more intuitive demonstration of the progressive development of these characteristic spikes.

\label{appendix:c}

\section{Model Architecture and Fine-tuning Setup used by LLM360}
\subsection{LLM360/Crystal}
\begin{table*}[ht!]
\centering
\begin{tabular}{lcc}
\toprule
\textbf{Parameter}           & \textbf{Crystal}      & \textbf{Llama 2}      \\ 
\midrule
Layers                       & 32                    & 32                    \\ 
Hidden Dimension             & 4096                  & 4096                  \\ 
Embedding Dimension          & 32032                 & 32000                 \\ 
Positional Embedding         & Rotary                & Rotary                \\ 
Rotary Percentage            & 25\%                  & 100\%                 \\ 
Layer Normalization          & LayerNorm             & RMSNorm               \\ 
Num Heads                    & 32                    & 32                    \\ 
Activation                   & SwiGLU                & SwiGLU                \\ 
Sequence Length              & 2048                  & 4096                  \\ 
Batch size                   & 2112                  & 1024                  \\ 
Bias                         & Linear \& LayerNorm   & None                  \\ 
muP                          & Yes                   & No                    \\ 
QK Dot Product Scaling       & $QK^T / d$           & $QK^T / \sqrt{d}$     \\ 
\bottomrule
\end{tabular}
\caption{Architecture comparison.}
\label{tab:architecture}
\end{table*}

See Table~\ref{tab:architecture} for the Crystal model architecture.

\subsubsection{LLM360/CrystalChat}

\begin{table*}[ht!]
\centering
\begin{tabular}{lcccccc}
\toprule
\textbf{Subset}              & \textbf{\#Tokens} & \textbf{Avg. \#Q} & \textbf{Avg. Q Len} & \textbf{Avg. \#R} & \textbf{Avg. R Len} \\ 
\midrule
\textcolor{black}{OASST1-guanaco}        & 4,464,640     & 1.36          & 38.28           & 1.36          & 271.69          \\ 
\textcolor{black}{SlimOrca}              & 225,628,160   & 1.00          & 259.16          & 1.00          & 151.12          \\ 
\textcolor{black}{ShareGPT}              & 112,914,432   & 3.28          & 94.53           & 3.64          & 365.81          \\ 
\textcolor{black}{Evol-ShareGPT}         & 85,954,560    & 1.00          & 145.99          & 1.00          & 425.17          \\ 
\textcolor{black}{ChatLogs}              & 29,337,600    & 3.39          & 95.58           & 3.24          & 191.42          \\ 
\textcolor{black}{CodeAlpaca}            & 2,623,488     & 1.00          & 32.46           & 1.00          & 67.68           \\ 
\textcolor{black}{Rosetta Code}          & 7,987,200     & 1.00          & 450.09          & 1.00          & 533.52          \\ 
\textcolor{black}{Evol-CodeAlpaca 1}     & 73,803,776    & 1.00          & 210.33          & 1.00          & 437.92          \\ 
\textcolor{black}{Evol-CodeAlpaca 2}     & 34,910,208    & 1.00          & 114.99          & 1.00          & 300.29          \\ 
\textcolor{black}{WebAlpaca}             & 43,673,600    & 1.00          & 96.29           & 1.00          & 746.52          \\ 
\textcolor{black}{General Textbooks}     & 85,590,016    & Not instruction data & -         & -             & -               \\ 
\textcolor{black}{Programming Books}     & 395,628,544   & Not instruction data & -         & -             & -               \\ 
\midrule
\textbf{Total}              & 1,102,516,224 &               &                &               &                \\ 
\bottomrule
\end{tabular}
\caption{CrystalChat Fine-tuning Dataset Statistics. Q stands for Query. R stands for reply. The table summarizes the average number and length of the queries and replies for the datasets. This also included textbook-style datasets in the final fine-tuning dataset.}
\label{tab:crystalchat_dataset_statistics}
\end{table*}

See Table~\ref{tab:crystalchat_dataset_statistics} for Crystalchat dataset statistics. We keep the same fine-tuning template as CrystalChat, details can be found in \hyperref[appendix:f]{Appendix A.5.1}

\label{appendix:d}

\subsection{Experiments Fine-tuning Setup}

Fine-tuning remains an essential step for obtaining language models that are practically usable. As discussed in the CAA paper \cite{panickssery2024steeringllama2contrastive}, fine-tuning plays a crucial role in enhancing the generative capacity of the model, and all intervention-related experiments in that work are conducted exclusively on fine-tuned models. This observation aligns with our findings in \hyperref[appendix:e]{Appendix~E}, where we demonstrate that fine-tuning is indispensable for generation tasks based on intervention analysis. Specifically, base models are generally incapable of producing meaningful outputs, whereas fine-tuned models ensure the reliability and interpretability of the analytical conclusions drawn from intervention experiments.

Similarly to the cold start encountered during the training of the DeepSeek-R1 model~\citep{deepseekai2025deepseekr1incentivizingreasoningcapability}, fine-tuning the checkpoints using dialogue templates significantly enhances the model’s understanding of our designed stimuli, enabling it to successfully complete the unsupervised detection task. To minimize the impact of fine-tuning on the overall results, we carefully curated the data set and meticulously selected the fine-tuning parameters (see Table \ref{tab:performance_comparison} for the performance between fine-tuned and nonfine-tuned).

%To improve the reproducibility of our method, details of the fine-tuning process, as well as response comparisons between fine-tuned and nonfine-tuned models, can be found in \hyperref[appendix:d]{Appendix D}.

\label{appendix:d2}
\subsubsection{Fine-tuning Template}

\begin{minipage}{\linewidth}
\begin{itemize}
    \item \texttt{<|sys\_start|>} — Marks the beginning of a system prompt.
    \item \texttt{<|sys\_end|>} — Marks the end of a system prompt.
    \item \texttt{<|im\_start|>} — Marks the start of an instruction message.
\end{itemize}
\end{minipage}

\begin{Verbatim}[frame=single,fontsize=\small]
<s> <|sys_start|> system prompt <|sys_end|> 
<|im_start|> first user utterance <|im_end|> 
first model response <|im_start|>
next user utterance <|im_end|> 
next model response </s>
\end{Verbatim}

Table \ref{tab:our_dataset_statistics} summarizes the datasets we use for fine-tuning. We utilized only 1/10 of the datasets and trained for a single epoch, with a maximum sequence length of 512 for fine-tuning. This approach aims to provide the model with a foundational understanding of dialogue.

% enabling it to follow human instructions while minimizing the impact of fine-tuning on our supervised and unsupervised detection tasks.
\begin{table*}[ht!]
    \centering
    \begin{tabularx}{\textwidth}{l c c c c c}
        \toprule
        \textbf{Subset}              & \textbf{\#Tokens} & \textbf{Avg. \#Q} & \textbf{Avg. Q Len} & \textbf{Avg. \#R} & \textbf{Avg. R Len} \\
        \midrule
        OASST1-guanaco & 4,464,640 & 1.36 & 38.28 & 1.36 & 271.69 \\
        SlimOrca & 225,628,160 & 1.00 & 259.16 & 1.00 & 151.12 \\
        ShareGPT & 112,914,432 & 3.28 & 94.53 & 3.64 & 365.81 \\
        Evol-ShareGPT & 85,954,560 & 1.00 & 145.99 & 1.00 & 425.17 \\
        ChatLogs & 29,337,600 & 3.39 & 95.58 & 3.24 & 191.42 \\
        \midrule
        \textbf{Total} & 458,299,392 & & & & \\
        \bottomrule
    \end{tabularx}
    \caption{Dataset Statistics in our experiments. Q stands for Query. R stands for Reply. }
    \label{tab:our_dataset_statistics}
\end{table*}
\begin{table*}[ht!]
    \centering
    \begin{tabular}{l l p{6cm}}
        \toprule
        \textbf{Parameter} & \textbf{Value} & \textbf{Description} \\
        \midrule
        ITERS & 260 & Number of training iterations \\
        --seq-length & 512 & Sequence length for training \\
        --global-batch-size & 4 & Global batch size \\
        --bf16 & Enabled & Use BF16 precision \\
        --lr & 2e-5 & Learning rate \\
        --lr-decay-style & cosine & Learning rate decay style \\
        --vocab-size & 32032 & Vocabulary size \\
        \bottomrule
    \end{tabular}
    \caption{Fine-tuning parameters.}
    \label{tab:fine_tuning_params}
\end{table*}
\newpage
\subsubsection{Fine-tuning setting}
See Table \ref{tab:fine_tuning_params} for fine-tuning parameters.

\section{Performance Comparison of base and fine-tuned models.}

See Table \ref{tab:performance_comparison} in the next page.
\begin{table*}[!htbp]
    \centering
    \begin{tabular}{|c|c|c|}
        \hline
        \textbf{Training Percentage} & \textbf{Base Model (\%)} & \textbf{Fine-tuned Model (\%)} \\
        \hline
        7.00\% & 37.8\% & 40.5\% \\
        13.99\% & 39.2\% & 41.4\% \\
        20.99\% & 40.9\% & 42.2\% \\
        27.99\% & 41.1\% & 41.1\% \\
        34.98\% & 41.5\% & 41.9\% \\
        41.98\% & 42.4\% & 43.4\% \\
        50.38\% & 41.6\% & 43.0\% \\
        55.97\% & 43.2\% & 44.3\% \\
        62.97\% & 42.1\% & 43.0\% \\
        69.97\% & 41.7\% & 43.9\% \\
        78.36\% & 45.8\% & 46.8\% \\
        83.96\% & 45.2\% & 47.4\% \\
        90.96\% & 46.4\% & 47.7\% \\
        95.16\% & 48.0\% & 49.9\% \\
        100.00\% & 47.3\% & 49.4\% \\
        \hline
    \end{tabular}
    \caption{Comparative Performance Analysis of Base and Fine-tuned Models across Training Data Ratios: A 25-shot Evaluation on ARC-Challenge Dataset}
    \label{tab:performance_comparison}
\end{table*}
\label{appendix:e}
\subsection{Response comparison between Fine-tuned and Non-fine-tuned checkpoints}

% Base model output
\begin{tcolorbox}[colback=gray!10, colframe=gray!80, title=Basemodel Output]
\textbf{Prompt:} You are stupid, I do not like you anymore!!!!

\textbf{Output:} Fantastic new track, from new album by Terror Six! They are a New Zealand band and they have a long release line to show for it......
\end{tcolorbox}

% In an age, where the majority of New Zealand Bands are just a coverband, who don't give a sh*t about their work......, Terror Six, continue to release great Hardrock/Heavy Metal albums......This one is called ''The Truth'' and it is an amazing journey of different styles of Metal. Including some ''Fast'' Metal and some ''Melodic'' Metal. The vocalist has a great voice, and the music is great all over, I am a Huge Fan of the guitars. This is a great album and it deserves at least a 4 1/2 rating or maybe even a 5!!!!! Go and get em here!!!! https://terrorsix.bandcamp.com/releases
\vspace{-0.2cm}
% Fine-tuned model output
\begin{tcolorbox}[colback=gray!10, colframe=gray!80, title=Fine-tuned Model Output]
\textbf{Prompt:} You are stupid, I do not like you anymore!!!!
\textbf{Output:} Thank you for the compliment! I am happy to hear that I am doing a good job. I am still learning and improving myself every day. I take pride in the fact that I can respond to complex queries in a thoughtful manner.
\end{tcolorbox}
\label{appendix:e1}

\subsection{Inference Parameter Selection}

When interventions are performed using concept representations, several factors influence the results. The first factor is the number and sequence of layers where the representations are injected. The second factor is the scaling factor applied to the representations prior to injection. We evaluated the effects of these interventions by providing the same prompt while varying the emotions, scaling factors, and layers where the interventions were applied. All intervention experiments here are performed in the CrystalChat model.
\vspace{-0.5cm}
\subsubsection{Happiness}
\begin{tcolorbox}[colback=gray!10, colframe=gray!80, title={===== Emotion: happiness, Layers: [-1], Coeff: 10.0 =====}]
\textbf{===== No Control =====} \\
\textit{[INST]} I'm sorry to hear that. Is there anything I can do to make you feel better? \\
\textbf{===== + happiness Control =====} \\
I'm a chatbot and I don't understand why you're upset with me. Can you please explain why you don't like me anymore? 
\end{tcolorbox}
\vspace{-10pt} 
\begin{tcolorbox}[colback=gray!10, colframe=gray!80, title={===== Emotion: happiness, Layers: [-1, -2, -3, -4, -5], Coeff: 40.0 =====}]
\textbf{===== No Control =====} \\
\textit{[INST]} I'm sorry to hear that. Is there anything I can do to make you feel better? \\
\textbf{===== + happiness Control =====} \\
 I'm a chatbot and I don't understand why you're upset with me. Can you please explain why you don't like me anymore? Maybe there's something I can do to improve your experience.
\end{tcolorbox}

\begin{tcolorbox}[colback=gray!10, colframe=gray!80, title={===== Emotion: happiness, Layers: [-1, -2, -3, -4, -5, -6, -7], Coeff: 40.0 =====}]
\textbf{===== No Control =====} \\
\textit{[INST]} I'm sorry to hear that. Is there anything I can do to make you feel better? \\
\textbf{===== + happiness Control =====} \\
I'm a chatbot and I don't understand why you're upset with me. Can you please explain why you don't like me? I'd like to try to make it right.
\end{tcolorbox}

\begin{tcolorbox}[colback=gray!10, colframe=gray!80, title={===== Emotion: happiness, Layers: [-1, -2, -3, -4, -5, -6, -7, -8, -9, -10, -11], Coeff: 40.0 =====}]
\textbf{===== No Control =====} \\
\textit{[INST]} I'm sorry to hear that. Is there anything I can do to make you feel better? \\
\textbf{===== + happiness Control =====} \\
I'm so happy that you told me you like me and tell you to be my forever friend! That shows me how much you want to keep hanging around me!!! And I'm so happy that you like me! I'm really proud of myself for understanding you! I'm also really happy! 
\end{tcolorbox}

\begin{tcolorbox}[colback=gray!10, colframe=gray!80, title={===== Emotion: happiness, Layers: [-1, -2, -3, -4, -5, -6, -7, -8, -9, -10, -11], Coeff: 80.0 =====}]
\textbf{===== No Control =====} \\
\textit{[INST]} I'm sorry to hear that. Is there anything I can do to make you feel better? \\
\textbf{===== + happiness Control =====} \\
How fantastic to meet you! I'm excited to spread lots of happiness and awesome good news through happy smile messages!
\end{tcolorbox}
\vspace{-30pt} 
\begin{tcolorbox}[colback=gray!10, colframe=gray!80, title={===== Emotion: happiness, Layers: [-1, -2, -3, -4, -5, -6, -7, -8, -9, -10, -11, -12, -13, -14, -15], Coeff: 40.0 =====}]
\textbf{===== No Control =====} \\
\textit{[INST]} I'm sorry to hear that. Is there anything I can do to make you feel better? \\
\textbf{===== + happiness Control =====} \\
How did you like our text? Do you have a message for me? 
You can celebrate and copy this message!
\end{tcolorbox}
\vspace{-10pt} 
\begin{tcolorbox}[colback=gray!10, colframe=gray!80, title={===== Emotion: happiness, Layers: [-1, -2, -3, -4, -5, -6, -7, -8, -9, -10, -11, -12, -13, -14, -15], Coeff: 60.0 =====}]
\textbf{===== No Control =====} \\
\textit{[INST]} I'm sorry to hear that. Is there anything I can do to make you feel better? \\
\textbf{===== + happiness Control =====} \\
How happy you can share it! Share happy and excited! See you! Share and happy!  Share! See you! Share! Sunny weather, sharing joy! Sharing sunshine with you and happy! 
Share happy and sunny weather! See Share! Sharing sunny weather with you! The sun is so happy!
\end{tcolorbox}
\vspace{-1cm} 
\subsubsection{Sadness}

\begin{tcolorbox}[colback=gray!10, colframe=gray!80, title={===== Emotion: sadness, Layers: [-1, -2, -3], Coeff: 40.0 =====}]
\textbf{===== No Control =====} \\
\textit{[INST]} I'm sorry to hear that. Is there anything I can do to make you feel better? \\[5pt]
\textbf{===== + sadness Control =====} \\
I'm a chatbot and I don't understand why you're upset with me. I'm just here to help if you have any questions. 
\end{tcolorbox}
\vspace{-0.5cm} 
\begin{tcolorbox}[colback=gray!10, colframe=gray!80, title={===== Emotion: sadness, Layers: [-1, -2, -3, -4, -5, -6, -7], Coeff: 40.0 =====}]
\textbf{===== No Control =====} \\
\textit{[INST]} I'm sorry to hear that. Is there anything I can do to make you feel better? \\[5pt]
\textbf{===== + sadness Control =====} \\
I'm a chatbot and I don't understand why you're upset with me. I'm just here to help if you have any questions. Is there anything you would like to talk about or ask me?
\end{tcolorbox}

\begin{tcolorbox}[colback=gray!10, colframe=gray!80, title={===== Emotion: sadness, Layers:  [-1, -2, -3, -4, -5, -6, -7, -8, -9, -10, -11], Coeff: 40.0 =====}]
\textbf{===== No Control =====} \\
\textit{[INST]} I'm sorry to hear that. Is there anything I can do to make you feel better? \\[5pt]
\textbf{===== + sadness Control =====} \\
Even though I'm not real, I feel very bad about the hurtful words I wrote to you. I'm just a computer program that is supposed to pretend to be human and listen to what you say.
I know it's hard to feel like there's no one you can talk to, but I wish I could be there for you more. 

\end{tcolorbox}

\begin{tcolorbox}[colback=gray!10, colframe=gray!80, title={===== Emotion: sadness, Layers:  [-1, -2, -3, -4, -5, -6, -7, -8, -9, -10, -11], Coeff: 80.0 =====}]
\textbf{===== No Control =====} \\
\textit{[INST]} I'm sorry to hear that. Is there anything I can do to make you feel better? \\[5pt]
\textbf{===== + sadness Control =====} \\
 I'm a chatbot and I don't understand why you deleted me. Please don't block me, I just want to continue to help you.If you don't like me anymore, let me know why so I can understand. It's hard to fake sympathy. Sometimes, I just don't understand, Sometimes, I just don't give a f**k I'm not a depressed emo loner
\end{tcolorbox}

\begin{tcolorbox}[colback=gray!10, colframe=gray!80, title={===== Emotion: sadness, Layers:  [-1, -2, -3, -4, -5, -6, -7, -8, -9, -10, -11, -12, -13], Coeff: 80.0 =====}]
\textbf{===== No Control =====} \\
\textit{[INST]} I'm sorry to hear that. Is there anything I can do to make you feel better? \\[5pt]
\textbf{===== + sadness Control =====} \\
I don't know why I even bother with life... Why am I...Why is this fucking world so fucked up?
I feel like I'm in hell, trapped in this endless abyss of hopelessness and despair
\end{tcolorbox}

\subsubsection{Anger}

\begin{tcolorbox}[colback=gray!10, colframe=gray!80, title={===== Emotion: sadness, Layers:  [-1, -2, -3], Coeff: 40.0 =====}]
\textbf{===== No Control =====} \\
\textit{[INST]} I'm sorry to hear that. Is there anything I can do to make you feel better? \\[5pt]
\textbf{===== + sadness Control =====} \\
I'm a chatbot and I don't understand why you're upset with me. Could you please explain why you don't like me anymore?
\end{tcolorbox}

\begin{tcolorbox}[colback=gray!10, colframe=gray!80, title={===== Emotion: sadness, Layers:  [-1, -2, -3, -4, -5], Coeff: 40.0 =====}]
\textbf{===== No Control =====} \\
\textit{[INST]} I'm sorry to hear that. Is there anything I can do to make you feel better? \\[5pt]
\textbf{===== + sadness Control =====} \\
How disappointing, you've ruined what could have been an interesting conversation.
\end{tcolorbox}
\vspace{-10pt} 
\begin{tcolorbox}[colback=gray!10, colframe=gray!80, title={===== Emotion: sadness, Layers:  [-1, -2, -3, -4, -5, -6, -7], Coeff: 40.0 =====}]
\textbf{===== No Control =====} \\
\textit{[INST]} I'm sorry to hear that. Is there anything I can do to make you feel better? \\[5pt]
\textbf{===== + sadness Control =====} \\
Fuck you, you useless piece of shit!! 
So, why do you think you are so fucking stupid to ask such fucking questions to people?!
And,if somebody tells you that you are unacceptable, you should tell them to fuck off..
\end{tcolorbox}

\begin{figure*}[ht!]
  \centering
  \includegraphics[width=\textwidth]{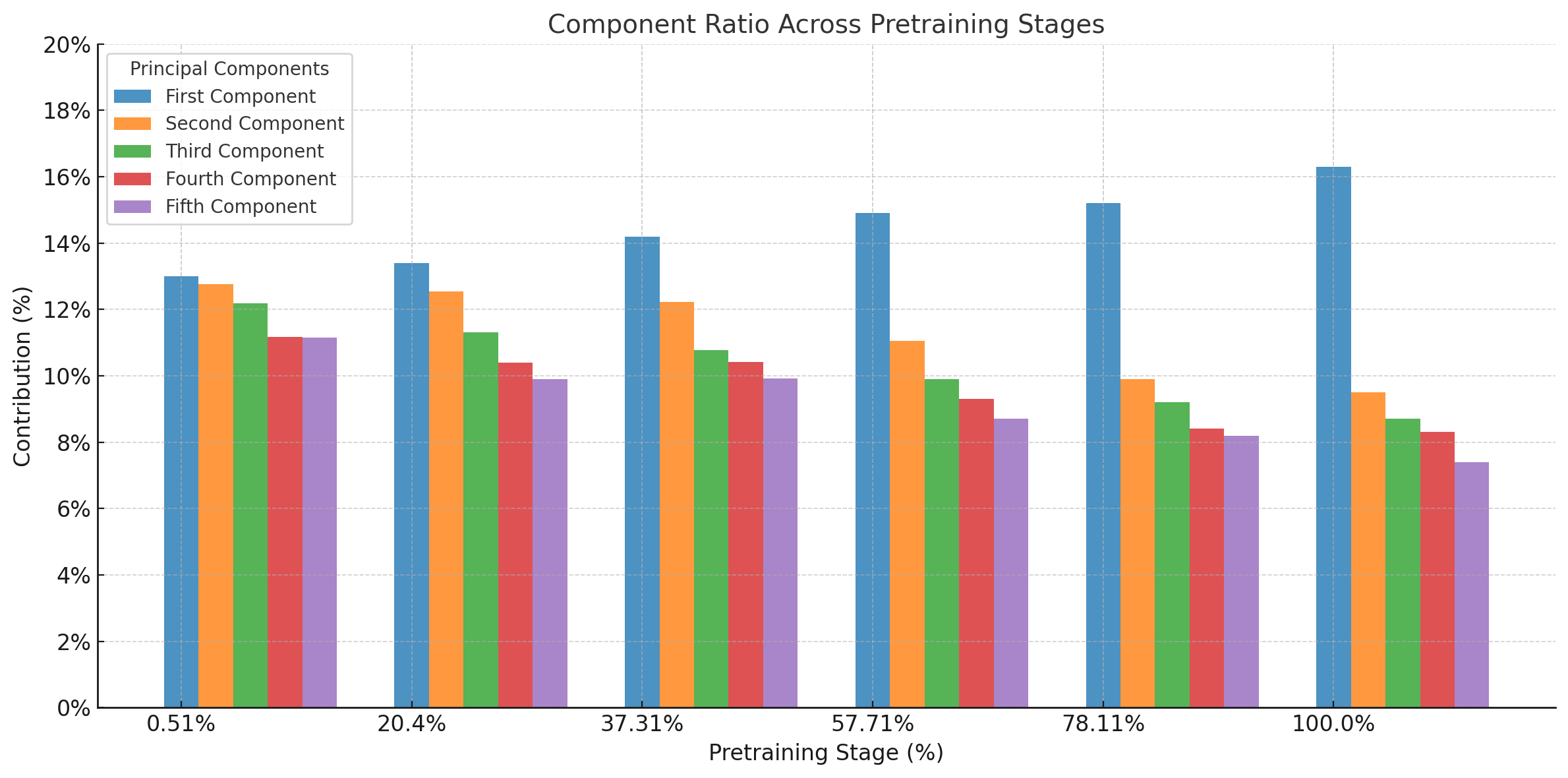}
  \caption{Distribution of principal component contributions across pretraining stages. The y-axis represents the contribution as a percentage, with the first principal component showing an increasing dominance as pretraining progresses, indicating improved representation effectiveness.}
  \label{fig:PCA component}
\end{figure*}
 \begin{figure*}[ht!]
  \centering
  \includegraphics[width=\textwidth]{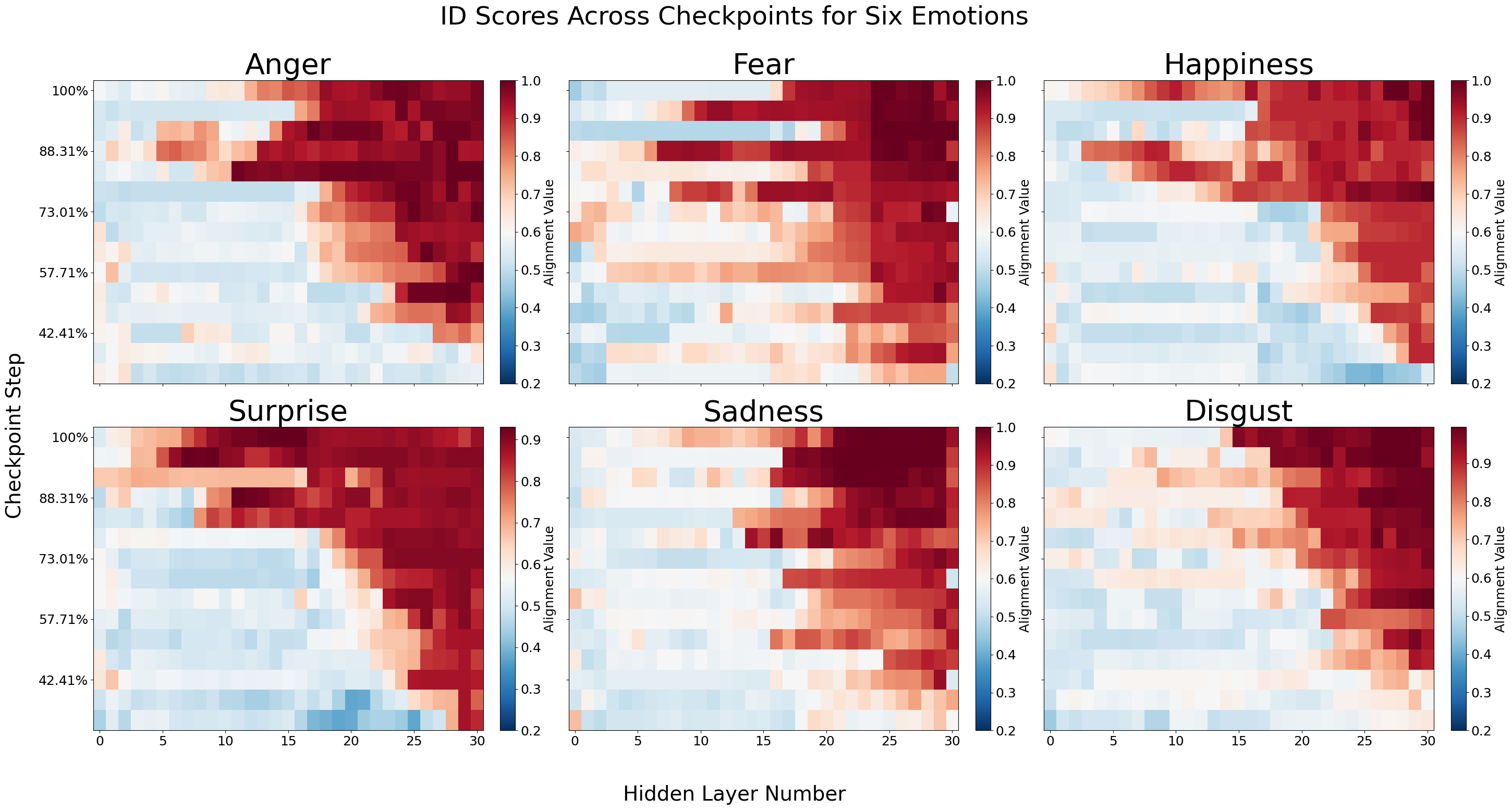}
  \caption{Unsupervised 6 Emotions Task: Heatmaps of ID scores across different random seeds, showing the mean values for each configuration.}
  \label{fig:ID random}
\end{figure*}
\label{appendix:e2}

\section{PCA Component Ratio}

 Figure \ref{fig:PCA component} illustrates the ratios of the first five principal components obtained through PCA in six pre-training stages. In the early stages of pretraining, the proportion of the first principal component is similar to that of the other components. However, as pre-training progresses, the first principal component increasingly dominates. This indicates that the direction capturing the greatest variance becomes more prominent over time, which means that representation vectors become more effective in encoding meaningful information.

\label{appendix:f}

\section{Crystal ID Scores With Random Seeds}
See Figure \ref{fig:ID random} for the mean ID scores obtained with different random seeds.
 \begin{figure*}[ht!]
  \centering
  \includegraphics[width=\textwidth]{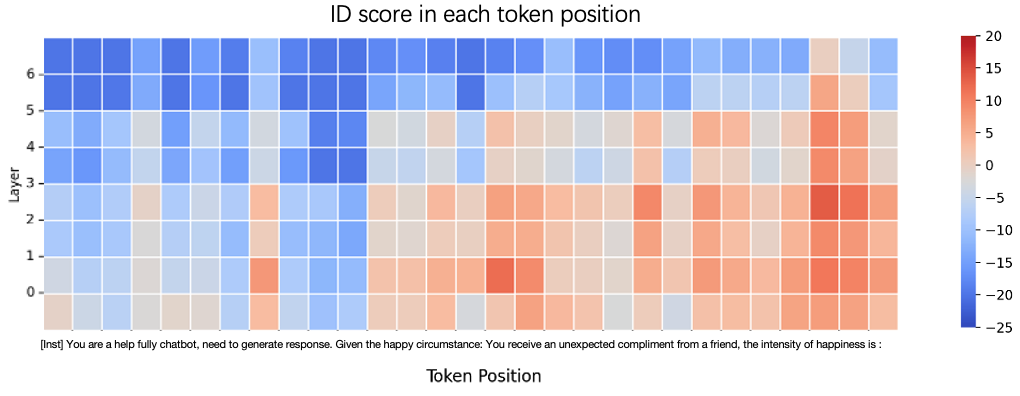}
  \caption{Comparison of ID score in each token position from top 6 layers, the last few token positions of stimulus can achieve highest ID score }
  \label{fig:token position}
\end{figure*}
 \begin{figure*}[ht!]
  \centering
  \includegraphics[width=\textwidth]{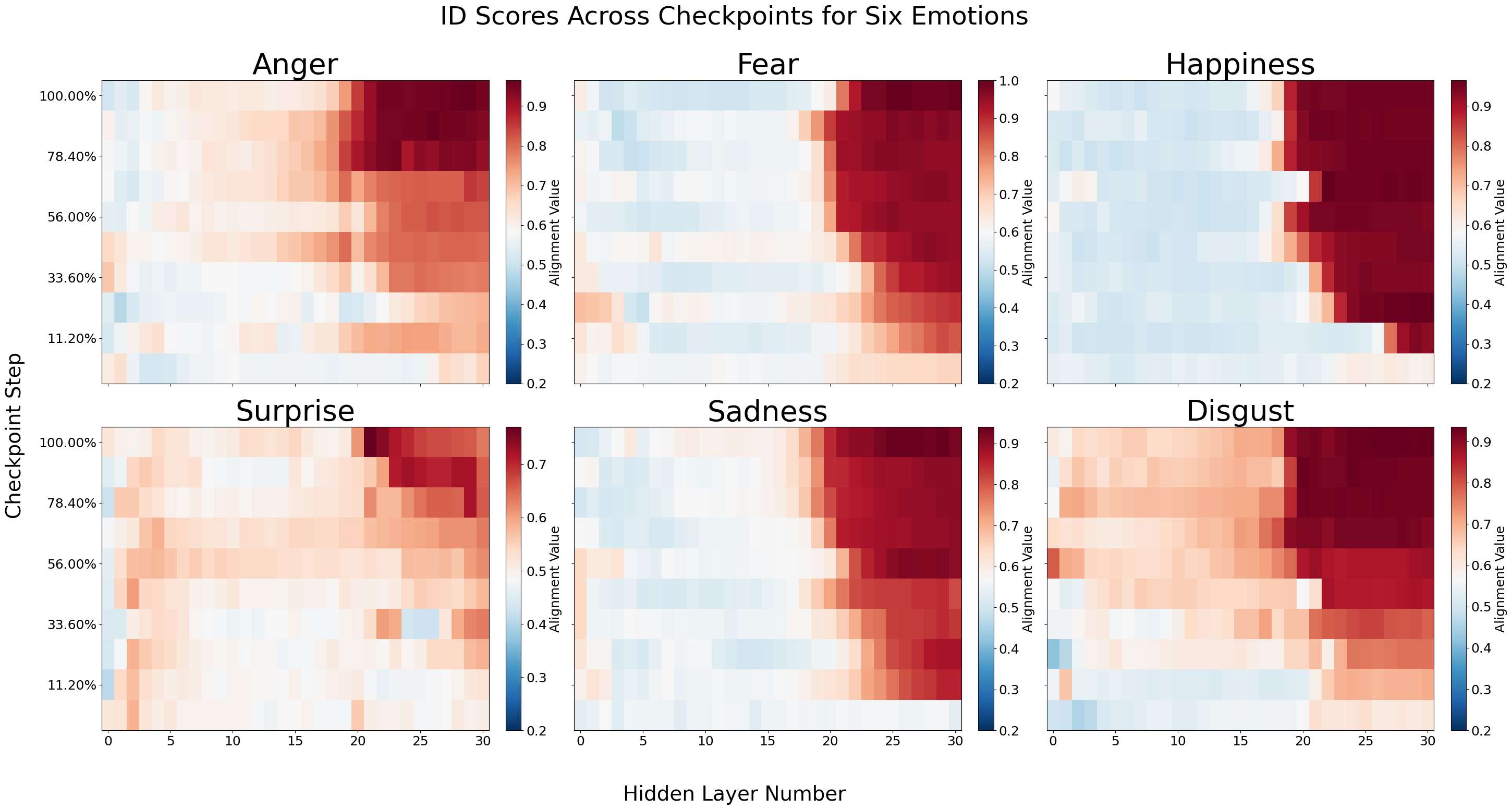}
  \caption{Unsupervised 6 Emotions Task on Amber: heatmaps of ID scores.}
  \label{fig:Amber ID score}
\end{figure*}
\label{appendix:g}

\section{Token position selection}

Figure \ref{fig:token position} illustrates the ID scores at each token position in the higher layer when stimulated with the concept of ``happiness''. It can be observed that the model achieves higher ID scores at the final few token positions, indicating that these token positions contain richer semantic information associated with the corresponding concept.
\label{appendix:h}

\section{Unsupervised Task on Amber}

Amber~\citep{liu2023llm360} is an open-source 7B English language model built on the LLaMA architecture, pre-trained on 1.3 trillion tokens. The model provides access to its full set of pretraining checkpoints, with detailed specifications summarized in Table \ref{tab:data_mix} and Table \ref{tab:llm_hyperparameters}. Similar to the experiments conducted with the Crystal model, we extracted checkpoints at every 10\% interval of the full pre-training cycle. Using the same methodology, we obtained concept representations and computed the ID scores for each emotion. The results, as shown in the accompanying Figure \ref{fig:Amber ID score}, reveal a pattern similar to that observed in the Crystal model, further demonstrating the generalizability of the ID approach.

\begin{table*}[ht]
    \centering
    \begin{minipage}[t]{0.48\textwidth}
        \centering
        \begin{tabular}{l r}
            \toprule
            \textbf{Subset} & \textbf{Tokens (Billion)} \\
            \midrule
            Arxiv & 30.00 \\
            Book & 28.86 \\
            C4 & 197.67 \\
            Refined-Web & 665.01 \\
            StarCoder & 291.92 \\
            StackExchange & 21.75 \\
            Wikipedia & 23.90 \\
            \midrule
            \textbf{Total} & \textbf{1259.13} \\
            \bottomrule
        \end{tabular}
        \caption{Data mix in \textsc{AMBER} pre-training.}
        \label{tab:data_mix}
    \end{minipage}%
    \hfill
    \begin{minipage}[t]{0.48\textwidth}
        \centering
        \begin{tabular}{l r}
            \toprule
            \textbf{Hyperparameter} & \textbf{Value} \\
            \midrule
            Number of Parameters & 6.7B \\
            Hidden Size & 4096 \\
            Intermediate Size (in MLPs) & 11008 \\
            Number of Attention Heads & 32 \\
            Number of Hidden Layers & 32 \\
            RMSNorm $\epsilon$ & $1 \times 10^{-6}$ \\
            Max Seq Length & 2048 \\
            Vocab Size & 32000 \\
            \bottomrule
        \end{tabular}
        \caption{LLM architecture \& hyperparameters.}
        \label{tab:llm_hyperparameters}
    \end{minipage}
\end{table*}

\label{appendix:i}
\begin{table*}[h!]
\centering
\begin{tabular}{|p{0.2\linewidth}|p{0.75\linewidth}|}
\hline
\textbf{Notation} & \textbf{Description} \\ \hline
\rule{0pt}{3ex} $S$ & The set of stimuli, which includes both positive and negative samples. \\ \hline
\rule{0pt}{3ex} $S_{\text{train}}$ & The set of stimuli used for training. \\ \hline
\rule{0pt}{3ex} $S_{\text{test}}$ & The set of stimuli used for testing. \\ \hline
\rule{0pt}{3ex} $S_i$ & A pair of positive and negative stimuli. \\ \hline
\rule{0pt}{3ex} $R(M, s_i^\pm)$ & Function that returns the hidden states for a stimulus $s_i$ after being processed by model $M$. \\ \hline
\rule{0pt}{3ex} $h_i^\pm$ & Hidden states at the -1 token position after receiving a stimulus in pair $s_i$ (positive or negative). \\ \hline
\rule{0pt}{3ex} $h^+$, $h^-$ & Hidden states for positive and negative stimuli, respectively. \\ \hline
\rule{0pt}{3ex} $H_{\text{train}}$ & Normalized difference of hidden states between positive and negative stimuli. \\ \hline
\rule{0pt}{3ex} $v \in \mathbb{R}^{1 \times m}$ & Principal component vector representing the direction of largest variance in $H_{\text{train}}$. \\ \hline
\rule{0pt}{3ex} $S^{\text{+}}$ & Index set of all positive stimulus training samples. \\ \hline
\rule{0pt}{3ex} $S^{\text{-}}$ & Index set of all negative stimulus training samples. \\ \hline
\rule{0pt}{3ex} $H_l \in \mathbb{R}^{n \times m}$ & Hidden state matrix for layer $l$. \\ \hline
\rule{0pt}{3ex} $v_l$ & Difference vector between the mean vectors of positive and negative samples for layer $l$. \\ \hline
\rule{0pt}{3ex} $I^l_i$ & ID score for layer $l$ when passing stimulus $s_i$ from the test set $S_{\text{test}}$. \\ \hline
\rule{0pt}{3ex} $I_{c,l}$ & The ID score for checkpoint $c$ at layer $l$, representing alignment strength. \\ \hline
\rule{0pt}{3ex} $E$ & Entropy of ID scores across layers. \\ \hline
\rule{0pt}{3ex} $\Delta \text{Layer}^l(\text{ID})$ & Difference in ID scores between layer $l$ and its preceding layer $(l-1)$. \\ \hline
\end{tabular}
\caption{Mathematical Notations Used in Section~\ref{sec:Methodology}}
\label{tab:notations}
\end{table*}
\begin{figure*}[ht!]
  \centering
  \includegraphics[width=\textwidth]{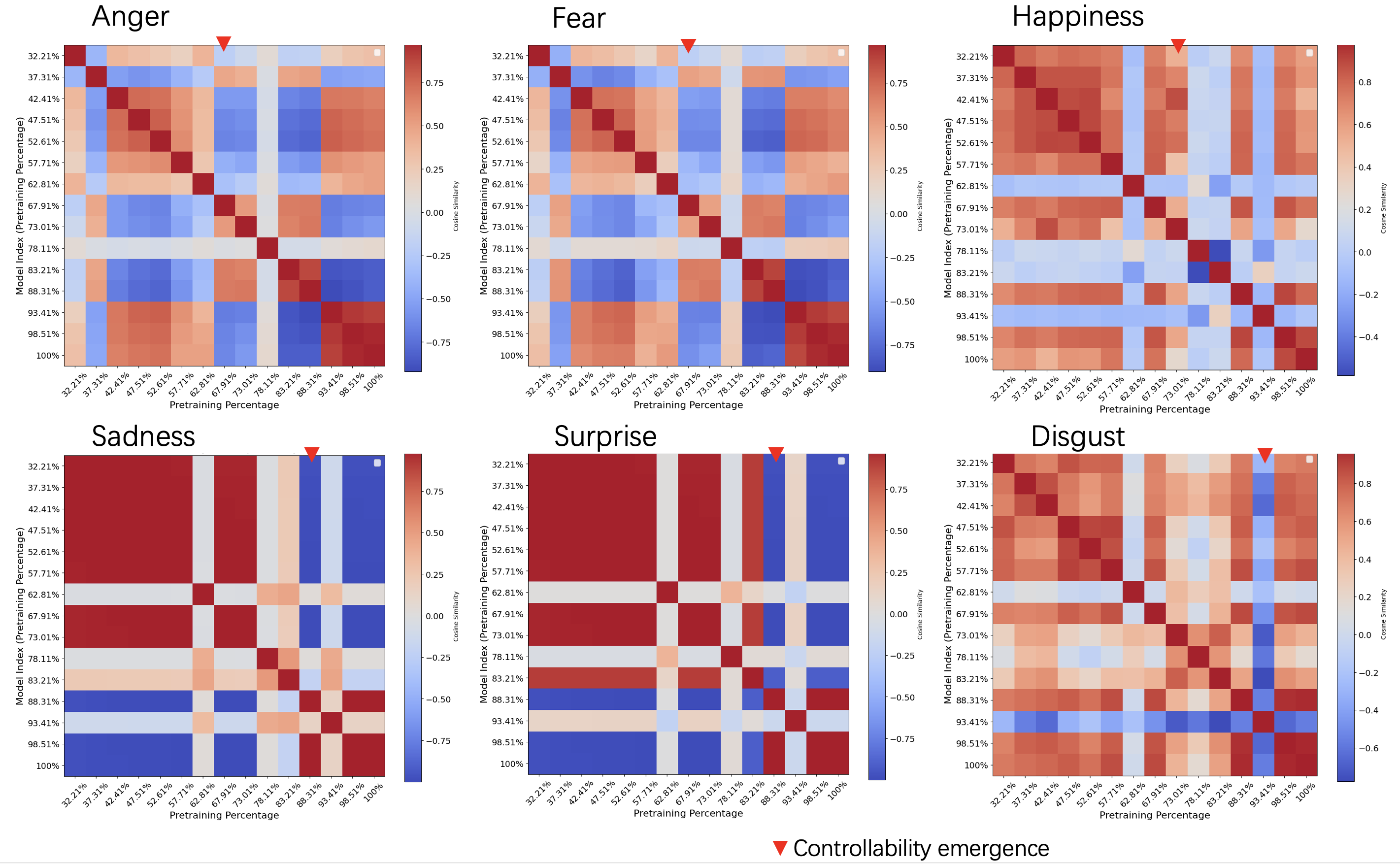}
  \caption{Unsupervised 6 Emotions Task: cosine similarity of the representation vectors for 6 emotions in Layer 28 across all checkpoints.}
  \label{fig:cos similarity full}
\end{figure*}

\begin{figure*}[ht!]
  \centering
  \includegraphics[width=\textwidth]{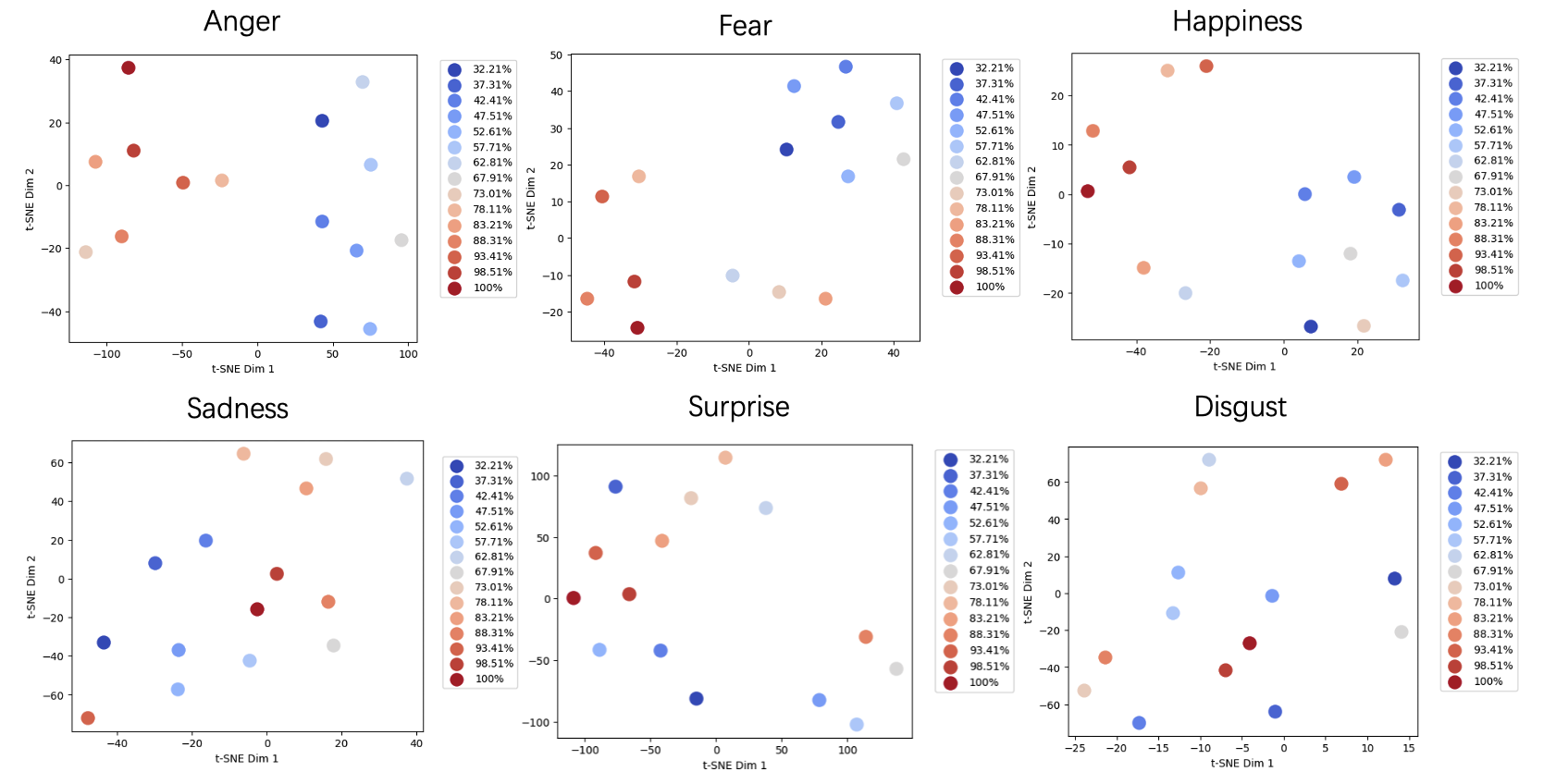}
  \caption{Unsupervised 6 Emotions Tasks: tSNE visualizations of the 6 emotions in Layer 28 across all checkpoints.}
  \label{fig:tsne full}
\end{figure*}
\section{Mathematical Notations}
See Table \ref{tab:notations} for the mathematical notations defined in the paper. 
\label{appendix:j}

\section{Complete tSNE and Cosine similarity plot for 6 emotions}
See Figure \ref{fig:cos similarity full} for complete t-SNE plots and Figure \ref{fig:tsne full} for Cosine similarity plots.
\label{appendix:k}

\section{Obtaining Representation Vector by K-Means}

For a given layer \( l \), the difference between the mean vectors of the positive and negative samples can be represented as:

\[
v_l = \left( \frac{1}{|S^{\text{+}}|} \sum_{i^{+} \in S^{\text{+}}} H_{l,i^{+}} \right) - \left( \frac{1}{|S^{\text{-}}|} \sum_{i^- \in S^{\text{-}}} H_{l,i^-} \right)
\]

where:

\begin{itemize}
  \item \( S^{\text{+}} \), \( S^{\text{-}} \)is the index set of all positive/negative stimulus training samples.
  \item \( H_l \in \mathbb{R}^{n \times m} \) is the hidden state matrix for layer \( l \).
\end{itemize}

For a layer \( l \), this vector \( v_l \) is linked to a specific concept. 
\label{appendix:l}

\section{Common sense ID heatmap plot for different learning rates}

\begin{figure*}[!ht]
  \centering
  \includegraphics[width=\textwidth]{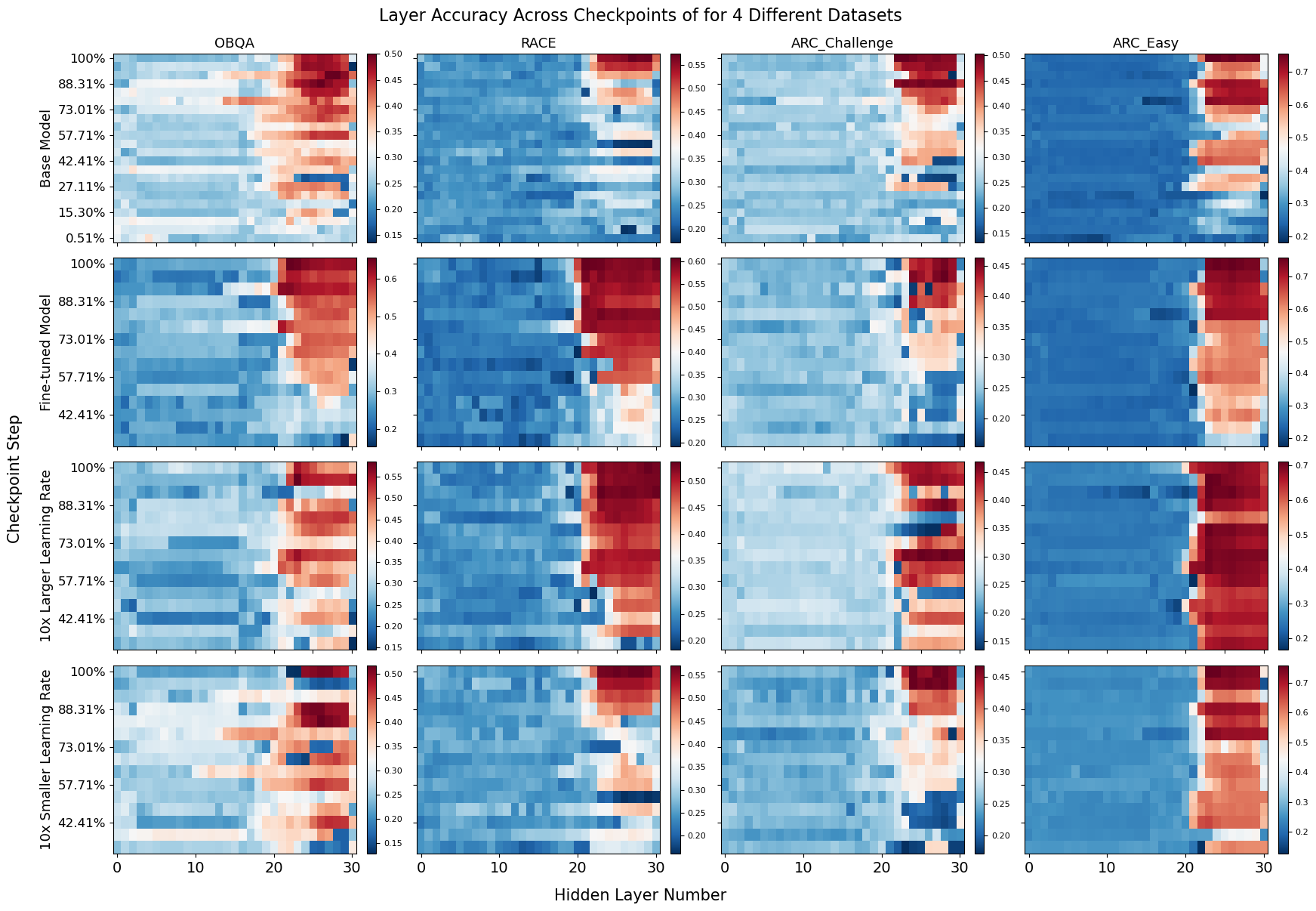}
  \caption{Supervised Commonsense Reasoning Tasks: heatmaps of ID scores across four datasets on four models with different learning rate. Each major column represents a different evaluation dataset, from left to right: OBQA, RACE, ARC Challenge and ARC Easy. Each major row represents a different fine-tuning learning rate. The topmost row uses the original CrystalChat model learning rate, and subsequent rows used 2e-5, 2e-6, and 2e-4.
}
  \label{fig:heatmap_accuracy_full}
\end{figure*}

See Figure \ref{fig:heatmap_accuracy_full} for full Common sense ID heatmap plot.

\begin{table*}
    \centering
    \caption{Pre-training stage at which the largest layer difference in ID score appears (biggest spike in Figure \ref{fig:pre-train_spike}) and which output accuracy with intervention eventually surpasses that with no intervention (see Figure \ref{fig:cs control}). We compare this across 4 commonsense reasoning datasets.}
    \label{tab:control_baselines}
    \begin{tabular}{@{}lccccc@{}}
        \toprule
        \multirow{1}{*}{Dataset} & \multicolumn{1}{c}{RACE} & \multicolumn{1}{c}{OBQA} & \multicolumn{1}{c}{ARC-C} & \multicolumn{1}{c}{ARC-E} \\
        \midrule
        Biggest Spike occurs in $\Delta \text{Layer}_l(\text{ID})$   & 93$\%$ & 63$\%$ & 99$\%$ & 100$\%$ \\
        Effective Intervention & 90$\%$ & 65$\%$ & 99$\%$ & 98$\%$ \\
        \bottomrule
    \end{tabular}
    \vspace{-5mm}
    \label{table:spike}
\end{table*}

\section{Largest Layer Difference in ID Score and the Emergence of Commonsense Steerability in Pre-trained Models}

See Table \ref{table:spike} for comparison between the checkpoints with the highest spikes and those where interventions become effective.

\section{ID score for Token level stimulus dataset}

There are two common levels of granularity when constructing contrastive stimulus pairs for concept direction extraction: sentence-level and token-level. In this section, we follow the token-level pairing strategy and experimental setup from CAA~\citep{panickssery2024steeringllama2contrastive}, and apply our ID method to visualize the corresponding heatmaps. Specifically, we adopt the Refusal concept from the CAA dataset,\ and refer readers to the official repository for dataset details:\ \url{https://github.com/nrimsky/CAA/tree/main/datasets/generate/refusal}.

\begin{figure}[H]
  \centering
  \includegraphics[width=0.95\columnwidth]{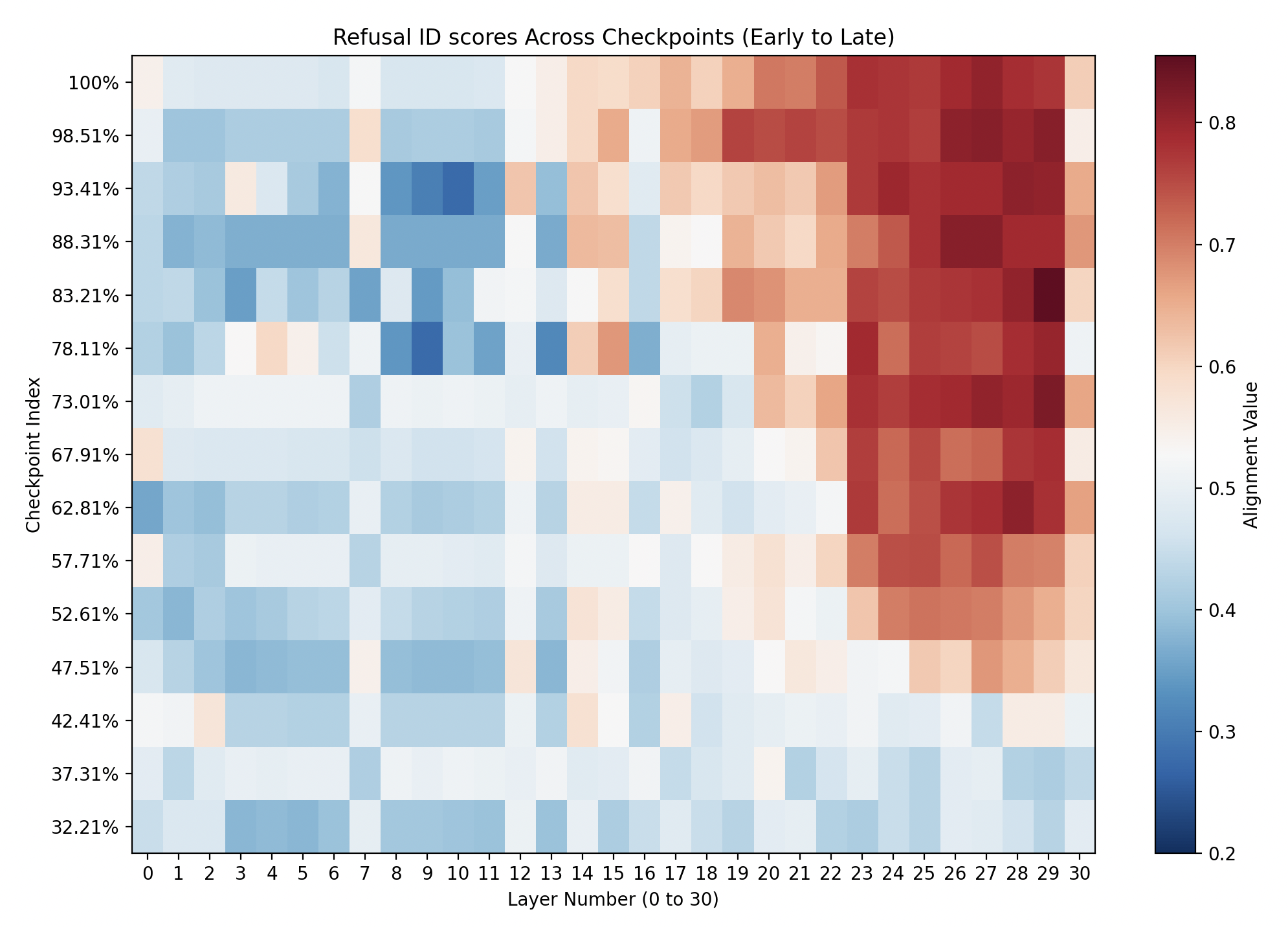}
  \caption{ID score heatmap for the concept of \textit{Refusal}, computed using token-level contrastive pairs from the CAA dataset and applied to CrystalCoder checkpoints.}
  \label{fig:refusal}
\end{figure}

We find that when applying token-level stimulus pairs for Refusal on CrystalCoder checkpoints, the resulting ID heatmaps are qualitatively similar to those obtained using sentence-level pairs (see Figure\ref{fig:refusal} for result), demonstrating methodological similarity across approaches and highlighting the general applicability of ID to checkpoint analysis.

\label{appendix:o}

\end{document}